\documentclass[11pt, a4paper]{lumia}

\usepackage[sort&compress]{natbib}
\usepackage{xspace}

\theoremstyle{plain}

\newtheorem*{proposition*}{Proposition}

\theoremstyle{definition}

\theoremstyle{definition}

\newcommand{\E}{\mathbb{E}}

\def\eqref#1{equation~\ref{#1}}

\usepackage{graphicx}           
\usepackage{tikz}               
\usepackage[edges]{forest}      

\usepackage{url}                
\usepackage{xurl}               

\usepackage{array}              
\usepackage{longtable}          
\usepackage{multirow}           
\usepackage{makecell}           
\usepackage{ragged2e}           

\usepackage{mathtools}          
\usepackage{nicefrac}           

\usepackage{algorithm}          
\usepackage{algorithmicx}       
\usepackage{algpseudocode}      
\usepackage{listings}           

\usepackage{subcaption}         
\usepackage{wrapfig}            
\usepackage[export]{adjustbox}  

\usepackage[commandnameprefix=always]{changes}            
\usepackage{xspace}             
\usepackage[normalem]{ulem}     
\usepackage{CJKutf8}            

\usepackage[tikz]{bclogo}       
\usepackage[framemethod=tikz]{mdframed} 

\usepackage{lipsum}             
\usepackage{tocloft}            
\usepackage{afterpage}          
\usepackage{bbding}             
\usepackage{epigraph}           
\usepackage{minitoc}            
\usepackage{multicol}           
\usepackage{textgreek}          
\usepackage[utf8]{inputenc} 
\usepackage[T1]{fontenc}    
\usepackage{hyperref}
\usepackage{url}
\usepackage{hyperref}       
\usepackage{url}            
\usepackage{booktabs}       
\usepackage{amsfonts}       
\usepackage{nicefrac}       
\usepackage{microtype}      
\usepackage{xcolor}         
\usepackage{wrapfig} 
\usepackage{amsmath,amssymb,amsfonts}
\usepackage{graphicx}
\usepackage{textcomp}
\usepackage{multirow} 
\usepackage{makecell}
\usepackage{authblk}
\usepackage{colortbl}
\usepackage{ulem}
\usepackage{wrapfig}
\usepackage{lipsum}  
\usepackage{tabularx}
\usepackage{adjustbox} 
\usepackage{enumitem}
\usepackage{nicefrac}       
\usepackage{etoolbox}
\usepackage{algorithm}      
\usepackage{algorithmicx}   
\usepackage{algpseudocode}  
\usepackage{subcaption}
\usepackage{caption}
\usepackage[skip=1pt]{caption}
\newcolumntype{P}[1]{>{\RaggedRight\arraybackslash}p{#1}}

\definecolor{uclablue}{RGB}{39, 116, 174}
\definecolor{bigaired}{RGB}{156, 0, 0}
\definecolor{myblue}{HTML}{598BE7}
\definecolor{mildblue}{RGB}{31,119,180}
\definecolor{sectionblue}{RGB}{70, 130, 180}
\definecolor{methodblue}{RGB}{0, 150, 136}
\definecolor{bgblue}{RGB}{245,243,253}
\definecolor{ttblue}{RGB}{91,194,224}
\definecolor{mygreen}{rgb}{0.64, 0.56, 0.88}
\definecolor{myyellow}{rgb}{0.68, 0.6, 0.1}
\definecolor{fancygreen}{rgb}{0.33, 0.68, 0.20}
\definecolor{salmon}{rgb}{0.94, 0.52, 0.49}
\definecolor{tablegreen}{rgb}{0.82, 0.94, 0.75}
\definecolor{tableblue}{rgb}{0.81, 0.90, 0.94}
\definecolor{tablered}{rgb}{0.97, 0.85, 0.85}
\definecolor{tableorange}{rgb}{0.96, 0.85, 0.81}
\definecolor{myorange}{rgb}{1.0, 0.49, 0.0}
\definecolor{tlgreen}{rgb}{0.33, 0.68, 0.20}
\definecolor{darkgreen}{RGB}{0,100,0}
\definecolor{darkred}{RGB}{200, 0, 0}
\definecolor{customyellow}{HTML}{FFFACD}
\definecolor{refinegreen}{RGB}{0, 128, 75}
\definecolor{scoregreen}{RGB}{34, 139, 34}
\definecolor{hidden-blue}{RGB}{194,232,247}
\definecolor{hidden-black}{RGB}{20,68,106}
\definecolor{yes}{HTML}{C6EFCE}
\definecolor{no}{HTML}{FFC7CE}
\definecolor{partial}{HTML}{FFEB9C}
\definecolor{external}{HTML}{D9E1F2}
\definecolor{hdr}{HTML}{F2F2F2}
\definecolor{GRPOrow}{gray}{0.96}
\definecolor{FlowRLrow}{RGB}{225,236,255}
\definecolor{FlowBlue}{RGB}{80,120,210}
\definecolor{GRPOGray}{gray}{0.35}

\hypersetup{
    colorlinks=true, 
    citecolor=uclablue, 
    linkcolor=bigaired,
    urlcolor=darkblue
}

\setlist[itemize]{leftmargin=20pt, noitemsep, topsep=0pt}

\NewDocumentCommand{\kaiyan}{mO{}}{\textcolor{purple}{\textsuperscript{\textit{kaiyan}}\textsf{\textbf{\small[#1]}}}}
\NewDocumentCommand{\yuxin}{mO{}}{\textcolor{cyan}{\textsuperscript{\textit{yuxin}}\textsf{\textbf{\small[#1]}}}}
\NewDocumentCommand{\bx}{mO{}}{\textcolor{green}{\textsuperscript{\textit{bx}}\textsf{\textbf{\small[#1]}}}}
\NewDocumentCommand{\at}{mO{}}{\textcolor{red}{\textsuperscript{\textit{AT}}\textsf{\textbf{\small[#1]}}}}
\NewDocumentCommand{\re}{mO{}}{\textcolor{blue}{\textsuperscript{\textit{RE}}\textsf{\textbf{\small[#1]}}}}
\NewDocumentCommand{\ybsun}{mO{}}{\textcolor{magenta}{\textsuperscript{\textit{youbang}}\textsf{\textbf{\small[#1]}}}}
\NewDocumentCommand{\runze}{mO{}}{\textcolor{orange}{\textsuperscript{\textit{runze}}\textsf{\textbf{\small[#1]}}}}
\NewDocumentCommand{\add}{mO{}}{\textcolor{darkgreen}{\textsuperscript{\textit{Maybe Consider Discuss}}\textsf{\textbf{[#1]}}}}

\newcommand{\cmark}{\textcolor{darkgreen}{\boldmath$\checkmark$}}
\newcommand{\xmark}{\textcolor{darkred}{\boldmath$\times$}}

\newenvironment{itemize*}%
 {\leftmargini=10pt\begin{itemize}%
  \setlength{\itemsep}{0pt}%
  \setlength{\parskip}{0pt}%
  }%
 {\end{itemize}}

\newenvironment{enumerate*}%
 {\begin{enumerate}%
  \setlength{\itemsep}{0pt}%
  \setlength{\parskip}{0pt}}%
 {\end{enumerate}}

\newcommand{\cellstatus}[1]{%
  \begingroup
  \StrTrim{#1}[\statusval]%
  \IfStrEq{\statusval}{Yes}{\cellcolor{yes}\cmark}{}%
  \IfStrEq{\statusval}{No}{\cellcolor{no}\xmark}{}%
  \IfBeginWith{\statusval}{Yes (}{\cellcolor{yes}\cmark~\textit{\statusval\unskip}}{}%
  \IfStrEq{\statusval}{Partial}{\cellcolor{partial}\textbf{Partial}}{}%
  \IfStrEq{\statusval}{External}{\cellcolor{external}\textbf{External}}{}%
  \endgroup
}

\newtcolorbox{myboxi}[1][]{
  breakable,
  title=#1,
  colback=red!5,
  colbacktitle=red!5,
  coltitle=black,
  fonttitle=\bfseries,
  bottomrule=0pt,
  toprule=0pt,
  leftrule=2pt,
  rightrule=2pt,
  titlerule=0pt,
  arc=0pt,
  outer arc=0pt,
  colframe=red,
}

\newtcolorbox{myboxnote}[1][]{
  breakable,
  title=#1,
  colback=orange!0,
  colbacktitle=orange!0,
  coltitle=black,
  fonttitle=\bfseries,
  bottomrule=0pt,
  toprule=0pt,
  leftrule=2pt,
  rightrule=2pt,
  titlerule=0pt,
  arc=0pt,
  outer arc=0pt,
  colframe=orange,
}

\newtcolorbox{myboxii}[1][]{
  breakable,
  freelance,
  title=#1,
  colback=white,
  colbacktitle=white,
  coltitle=black,
  fonttitle=\bfseries,
  bottomrule=0pt,
  boxrule=0pt,
  colframe=white,
  overlay unbroken and first={
  \draw[red!75!black,line width=3pt]
    ([xshift=5pt]frame.north west) -- 
    (frame.north west) -- 
    (frame.south west);
  \draw[red!75!black,line width=3pt]
    ([xshift=-5pt]frame.north east) -- 
    (frame.north east) -- 
    (frame.south east);
  },
  overlay unbroken app={
  \draw[red!75!black,line width=3pt,line cap=rect]
    (frame.south west) -- 
    ([xshift=5pt]frame.south west);
  \draw[red!75!black,line width=3pt,line cap=rect]
    (frame.south east) -- 
    ([xshift=-5pt]frame.south east);
  },
  overlay middle and last={
  \draw[red!75!black,line width=3pt]
    (frame.north west) -- 
    (frame.south west);
  \draw[red!75!black,line width=3pt]
    (frame.north east) -- 
    (frame.south east);
  },
  overlay last app={
  \draw[red!75!black,line width=3pt,line cap=rect]
    (frame.south west) --
    ([xshift=5pt]frame.south west);
  \draw[red!75!black,line width=3pt,line cap=rect]
    (frame.south east) --
    ([xshift=-5pt]frame.south east);
  },
}

\tcbset{
  takeawaysbox/.style={
    title=Takeaways,
    colback=lightblue!80,
    colframe=black,
    fonttitle=\bfseries\small,
    coltitle=white,
    colbacktitle=black,
    enhanced,
    attach boxed title to top left={xshift=2.5mm,yshift=-2.5mm},
    boxed title style={rounded corners, size=small, colframe=black, colback=black},
    width=\linewidth,
    arc=3.5mm
  }
}

\mdfdefinestyle{mystyle}{%
  rightline=true,
  innerleftmargin=10,
  innerrightmargin=10,
  outerlinewidth=3pt,
  topline=false,
  rightline=true,
  bottomline=false,
  skipabove=\topsep,
  skipbelow=\topsep
}

\tikzset{%
    every node/.style={font=\tiny},
    parent/.style =          {align=center,text width=2cm,rounded corners=3pt, line width=0.3mm, fill=gray!10,draw=gray!80},
    child/.style =           {align=center,text width=2.0cm,rounded corners=3pt, fill=blue!10,draw=blue!80,line width=0.3mm},
    grandchild/.style =      {align=center,text width=2cm,rounded corners=3pt},
    greatgrandchild/.style = {align=center,text width=1.5cm,rounded corners=3pt},
    greatgrandchild2/.style = {align=center,text width=1.5cm,rounded corners=3pt},    
    referenceblock/.style =  {align=center,text width=1.5cm,rounded corners=2pt},
    pretrain/.style =           {align=center,text width=2.0cm,rounded corners=3pt, fill=blue!10,draw=blue!80,line width=0.3mm},   
    pretrain_work/.style =           {align=center, text width=8.5cm,rounded corners=3pt, fill=blue!10,draw=blue!0,line width=0.3mm},  
    template/.style =           {align=center,text width=2.0cm,rounded corners=3pt, fill=red!10,draw=red!80,line width=0.3mm},   
    template_work/.style =           {align=center,text width=8.5cm,rounded corners=3pt, fill=red!10,draw=red!0,line width=0.3mm},    
    answer/.style =           {align=center,text width=2.0cm,rounded corners=3pt, fill= cyan!10,draw= cyan!80,line width=0.3mm},   
    answer_work/.style =           {align=center,text width=8.5cm,rounded corners=3pt, fill= cyan!10,draw= cyan!0,line width=0.3mm},      
    multiple/.style =           {align=center,text width=2.0cm,rounded corners=3pt, fill= orange!10,draw= orange!80,line width=0.3mm},   
    multiple_work/.style =           {align=center,text width=8.5cm,rounded corners=3pt, fill= orange!10,draw= orange!0,line width=0.3mm},        
    tuning/.style =           {align=center,text width=2.0cm,rounded corners=3pt, fill= magenta!10,draw= magenta!80,line width=0.3mm},   
    tuning_work/.style =           {align=center,text width=8.5cm,rounded corners=3pt, fill= magenta!10,draw= magenta!0,line width=0.3mm},          
}

\newcommand{\lstbg}[3][0pt]{{\fboxsep#1\colorbox{#2}{\strut #3}}}

\lstdefinelanguage{diff}{
  basicstyle=\ttfamily\small,
  morecomment=[f][\lstbg{red!20}]-,
  morecomment=[f][\lstbg{green!20}]+,
}

\lstdefinelanguage{diffpython}{
  language=diff,
  morekeywords={def, if, else, for, while, return, import, from, as, class, with, try, except, finally, raise, lambda, and, or, not, in, is, None, True, False},
  morecomment=[l]{\#},
  morestring=[b]",
  morestring=[b]',
}

\newcommand{\MethodName}{ContextLM}

\setheadertext{LUMIA Lab}

\correspondingemail{\emailicon beiya\_dai@sjtu.edu.cn \quad lin.zhouhan@gmail.com \quad $^\ddagger$ Corresponding Author. }

\githublink{https://github.com/LUMIA-Group/ContextLM}

\huggingfacelink{https://huggingface.co/daibeiya/ContextLM}

\renewcommand{\today}{2025-10-01}

\title{Context-level Language Modeling \\ by Learning Predictive Context Embeddings}
\setheadertitle{Context-level Language Modeling by Learning Predictive Context Embeddings}

\renewcommand\Authfont{\normalfont\bfseries\fontsize{9}{13}\selectfont}
\renewcommand\Affilfont{\normalfont\fontsize{8}{10}\selectfont}
\author{%
    \textbf{Beiya Dai}$^{1}$, \textbf{Yuliang Liu}$^{1,5}$, \textbf{Daozheng Xue}$^{1}$, \textbf{Yunchong Song}$^{2}$, \textbf{Qipeng Guo}$^{2}$, \textbf{Kai Chen}$^{2}$, \textbf{Xinbing Wang}$^3$,  \\
    \Authfont \textbf{Bowen Zhou}$^{2,4}$, \textbf{Zhouhan Lin}$^{1,2 \ddagger}$ \\
    $^1$ LUMIA Lab, School of Artificial Intelligence, Shanghai Jiao Tong University\\
    $^2$ Shanghai AI Laboratory \\
    $^3$ Shanghai Jiao Tong University\quad
    $^4$ Tsinghua University\quad 
    $^5$ Nanjing University \\
}

\begin{document}

\begin{abstract}
We propose ContextLM, a framework that implicitly learns multi-token prediction by augmenting standard pretraining with an intrinsic next-context prediction objective. ContextLM builds a language model on top of context embeddings that span multiple tokens, enabling better next-token prediction by predicting the next context. Our model is fully compatible with standard autoregressive, token-by-token evaluation paradigms (e.g., perplexity). Extensive experiments with GPT-2 and Pythia backbones (up to 1.5B parameters and 300B training tokens) reveal that ContextLM shifts the Pareto frontier of scaling laws, exhibiting superior efficiency in parameters, training tokens, and FLOPs. Our results show that ContextLM could already achieve the baseline perplexity using 39\% fewer parameters and demonstrates robust generalization improvements on extensive downstream tasks under equivalent parameter counts.
\end{abstract}
 
\maketitle

\begin{figure*}[hb]
      \centering
      \begin{subfigure}{0.328\textwidth}
        \includegraphics[width=\linewidth]{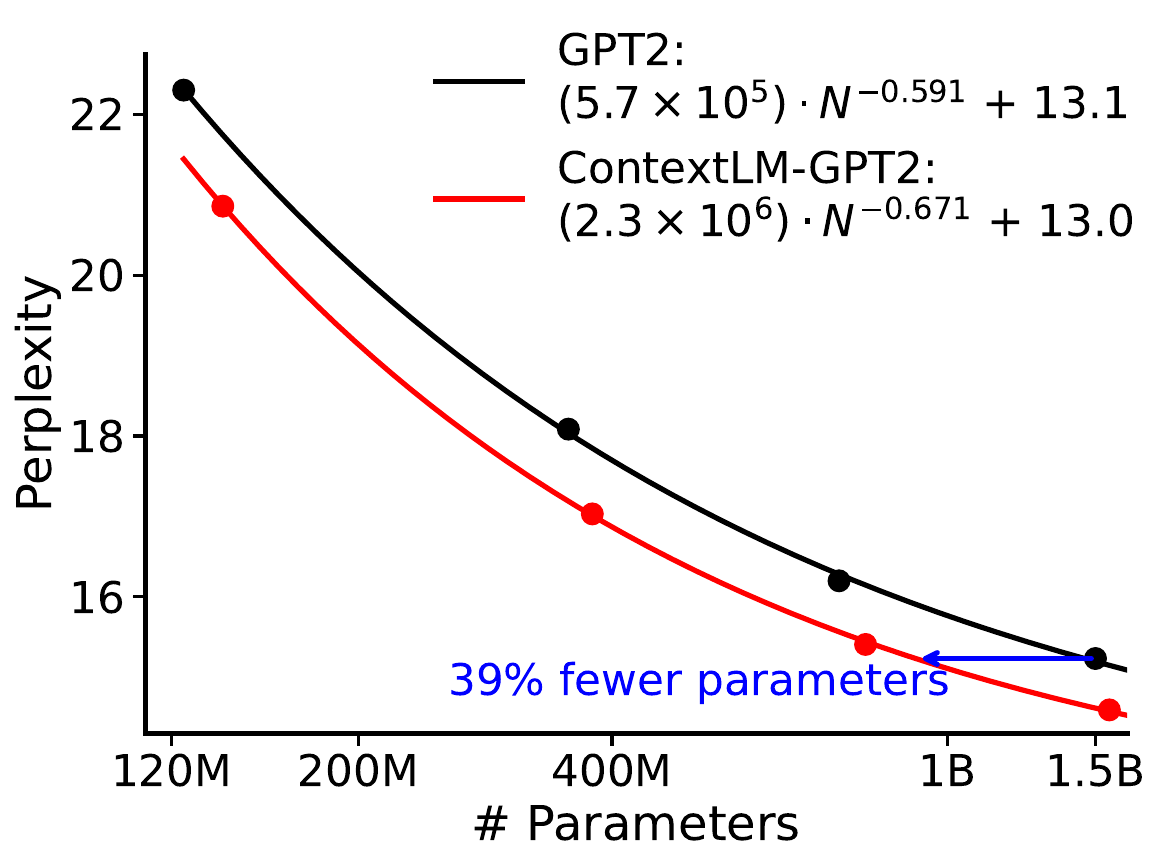}
        \subcaption{Parameters (log scale)}
        \label{fig:scaling_law_parameters}
      \end{subfigure}
      \begin{subfigure}{0.328\textwidth}
          \includegraphics[width=\linewidth]{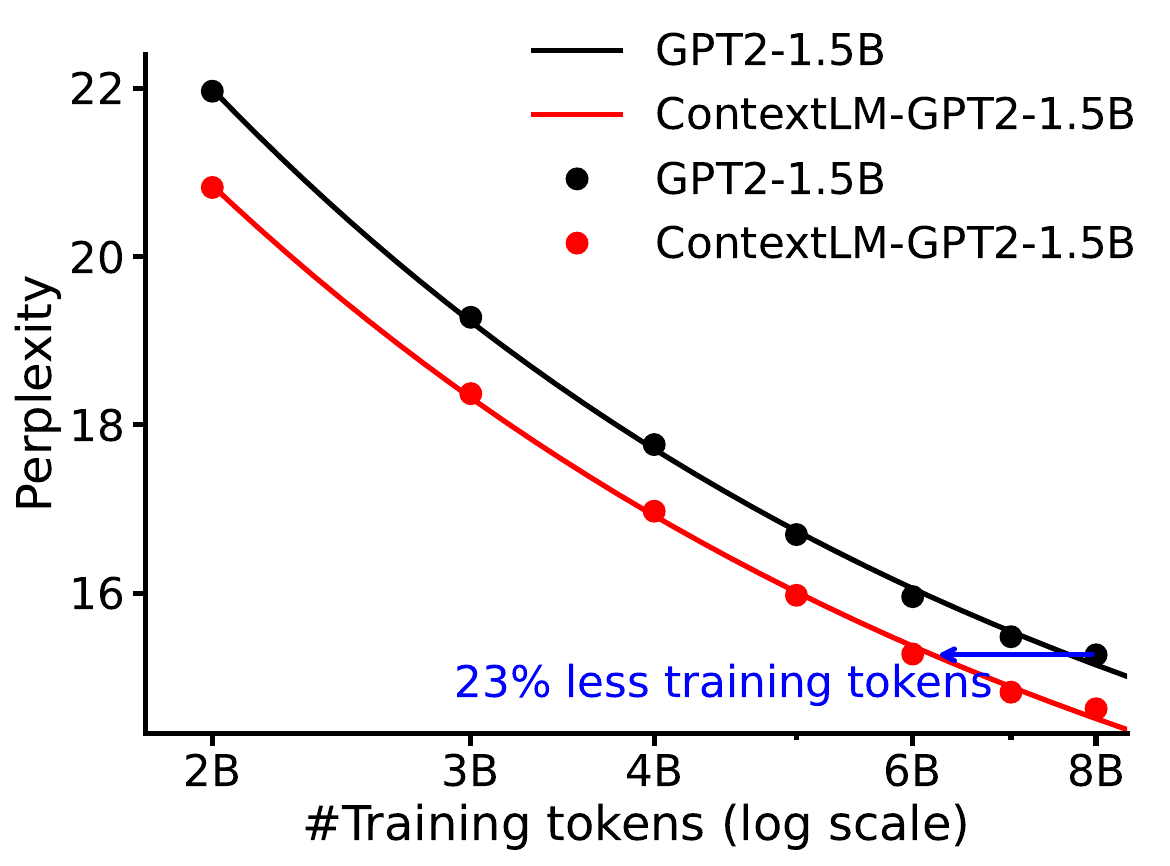}
        \subcaption{Training Tokens (log scale)}
        \label{fig:scaling_law_data}
      \end{subfigure}
      \begin{subfigure}{0.328\textwidth}
          \includegraphics[width=\linewidth]{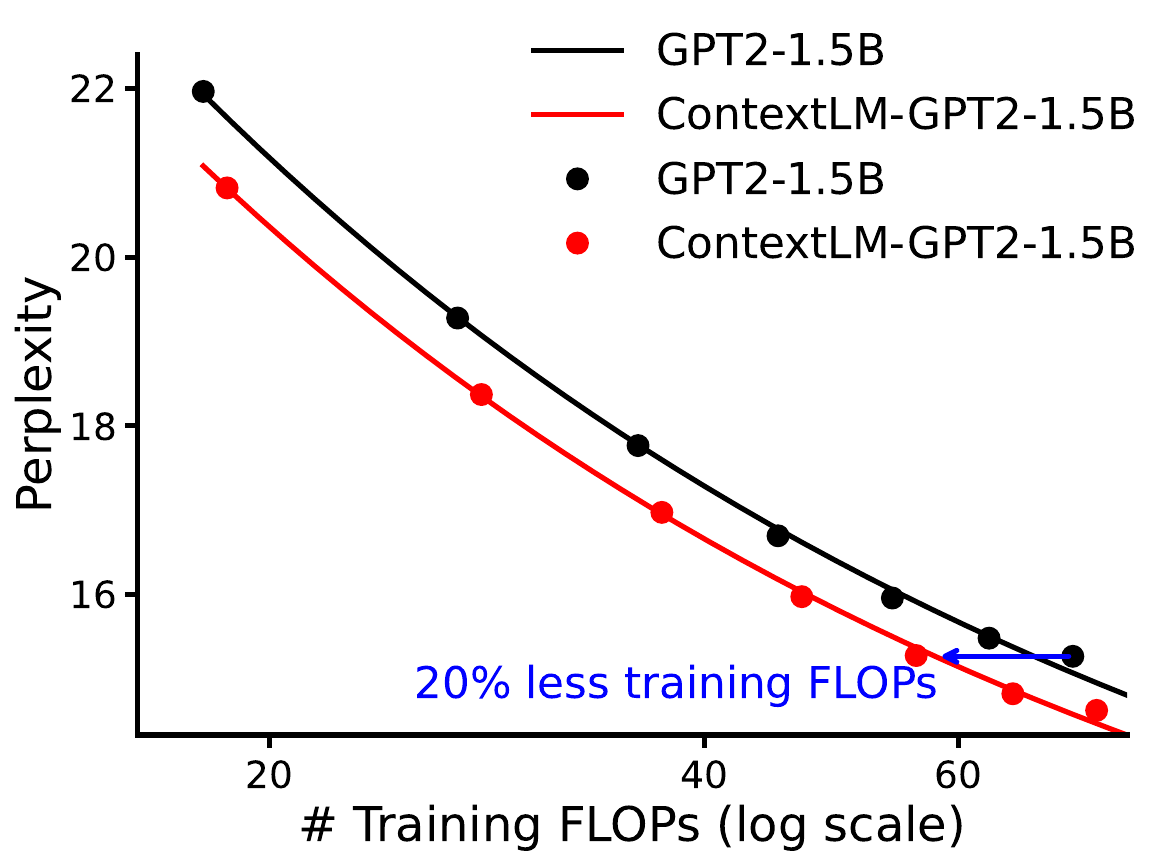}
            \subcaption{Training FLOPs (log scale)}
        \label{fig:scaling_law_flops}
      \end{subfigure}
      \caption{
      Scaling performance comparison across three dimensions: parameters, training tokens, and training FLOPs for GPT-2 and \MethodName-GPT2. 
      }
        \label{fig:scaling_law}
\end{figure*}
 
\section{Introduction}
Large language models (LLMs) have become the foundation of modern natural language processing (NLP), demonstrating remarkable capabilities in text generation, logical reasoning, and generalization ability~\citep{radford2019language,brown2020language,Achiam2023GPT4TR,touvron2023llama}. Historically, language models have mainly been explored at two levels of granularity: character-level~\citep{graves2013generating,kim2016character} and token-level~\citep{bengio2003neural}. Modern LLMs have universally converged on the token-level paradigm, as token-level prediction provides a more challenging training objective. 
This dominance of token-level models stems from its requirement to predict over more sophisticated dynamics that transcend character combinations, thereby yielding more powerful models.

Recent works have explored more challenging pretraining tasks, such as multi-token prediction (MTP), which extends next-token prediction (NTP) to predict multiple future tokens~\citep{gloeckle2024better, shao2024beyond}.
Although MTP advances beyond token-level prediction via its training objective, the model's hidden representations remain at the token level. Moving beyond such token-level representations requires fundamental architectural innovations. 
The pivotal challenge here lies in the construction and utilization of abstract-level representations. The pioneering framework JEPA~\citep{lecun2022jepa} addresses this by introducing a systematic approach to learning multi-level representations, shifting the predictive objective to a latent representation space. This framework has been successfully applied in the image and video domains~\citep{bardes2023v,tuncay2025audio}.

For natural language, efforts to introduce hierarchical abstraction have been made primarily through two directions. Sentence-level abstractions like LCM~\citep{lcmteam2024largeconceptmodelslanguage} and Block Transformer~\citep{ho2024block} successfully compress information at the cost of sacrificing autoregressive modeling in the latent space.
On the other hand, character-level methods such as BLT~\citep{pagnoni2024byte} and H-Net~\citep{Hwang2025DynamicCF} construct latent token-level representations by dynamically grouping characters into patches.
While these models can learn to form dynamic token-level representations through training, their effectiveness to learn representations beyond the token-level still remains challenging.





In this work, we propose \textbf{ContextLM}, a framework that introduces an extra intrinsic context-level prediction mechanism that operates in the latent space.
Our approach decomposes the Transformer into a Token Encoder and a Token Decoder, with an additional Context Predictor in between. This predictor operates autoregressively in latent space, generating predictive context embeddings to semantically guide the token generation. Instead of explicitly predicting discrete tokens, ContextLM employs an intrinsic next-context prediction objective for the Context Predictor, optimizing contexts to minimize aggregated signals from future tokens. 
Crucially, ContextLM maintains full architectural compatibility with standard Transformers, allowing seamless integration with existing mainstream LLM evaluation metrics such as token-level perplexity.

Our empirical evaluation demonstrates that ContextLM significantly shifts the Pareto frontier of efficient language modeling. Using GPT-2 backbones, ContextLM matches baseline perplexity using 39\% fewer parameters, 23\% fewer training tokens, and 20\% fewer FLOPs (see Figure~\ref{fig:scaling_law}). We further validate the scalability of our approach by training a ContextLM variant based on the Pythia-1.4B architecture on 300B tokens. The results show that ContextLM not only improves perplexity but also translates these gains into superior downstream task performance and stronger instruction-following capabilities, suggesting that latent context prediction offers a promising path toward stronger language models.

    \begin{figure*}[ht]
        \centering
        \scalebox{1}{
      \begin{subfigure}{0.52\textwidth}
        \includegraphics[width=\linewidth]{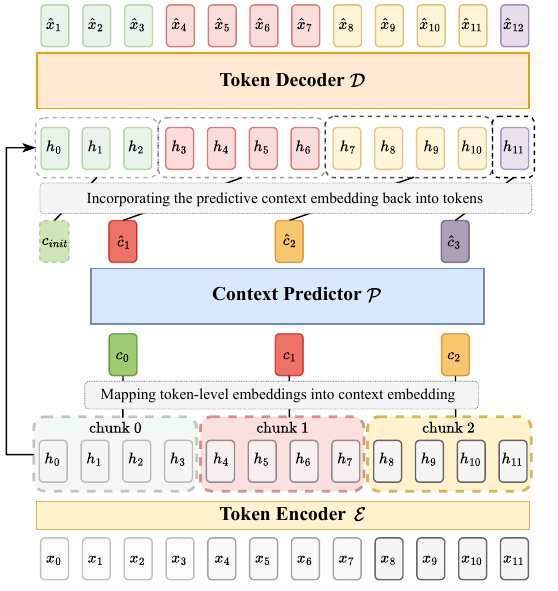}
        \end{subfigure}
      \hspace{8pt}
      \begin{subfigure}{0.43\textwidth}
        \includegraphics[width=\linewidth]{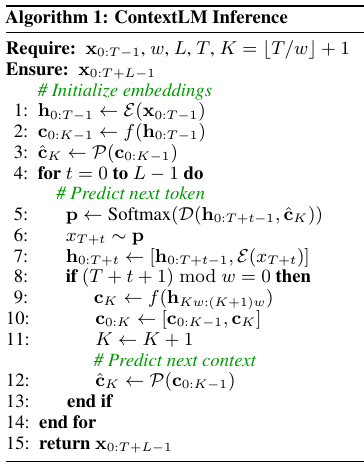}
        \end{subfigure}}
    \caption{Overview of the ContextLM architecture. 
    The model extends the standard language model by introducing a Context Predictor $\mathcal{P}$ 
    which generates context embeddings that are fed into the Token Decoder $\mathcal{D}$ to guide token generation. 
    The pseudocode details the inference procedure, showing the integration of context prediction into the token generation loop. Notably, the context prediction step only occurs at chunk boundaries.}
    \label{fig:contextLM}
    \end{figure*}
    \section{\MethodName}
    In this section, we present \MethodName\hspace{1pt}, 
    which constructs predictive context embeddings optimized by aggregated multi-token error signals, thereby augmenting conventional next-token modeling.
    We first introduce the problem setup, followed by the model architecture and training objective. In addition, we provide an analysis of the computational complexity in Sec.~\ref{sec: computational complexity}.

    \subsection{Problem Setup}
    \label{sec: Problem Setup}
    Given a text corpus $\mathcal{C}$, the standard language modeling objective is to model sequence $\mathbf{x}_{0:T-1} = (x_0, x_1, \ldots, x_{T-1})$, where $\mathbf{x}_{0:T-1} \in \mathcal{C}$, using a unidirectional autoregressive model $\theta_0$. This standard language model estimates token probabilities $p_{\theta_0}(x_t \mid x_{<t})$ based solely on preceding tokens. 

    We aim to augment the standard language model with latent, context-level representations that encode semantic dependencies beyond adjacent tokens. 
    To this end, we introduce a hierarchical architecture comprising three components: a {Token Encoder} to extract token-level representations, a {Context Predictor} to model the evolution of high-level semantics across chunks, and a {Token Decoder} to generate text conditioned on both token-level and predicted context embeddings.
    
    The Context Predictor is trained to autoregressively predict the embedding of the next context. The resulting model defines a context-aware language model $\theta$ that conditions token generation on both previous tokens and predicted context embeddings: $\theta(x_t \mid x_{<t}, \hat{c}_k)$, where $k = \lfloor t / w \rfloor+1$ denotes the current chunk index and $w$ is the chunk size. Here, $\hat{c}_k$ is the predicted context embedding corresponding to the current chunk.
     
    \subsection{Model Architecture}
    \label{sec: architecture}
    The overall architecture of \MethodName \hspace{1pt}, illustrated in Figure~\ref{fig:contextLM}, extends standard NTP with a context prediction pathway to capture multi-granularity semantic dependencies. It consists of three principal components: Token Encoder, Context Predictor, and Token Decoder.
    
        \subsubsection{Token Encoder $\mathcal{E}$}
        The Token Encoder $\mathcal{E}$ follows a standard decoder-only Transformer design and operates at the token level. Given an input token sequence $\mathbf{x}_{0:T-1}$, it produces token-level embeddings:
        \begin{equation}
            \mathbf{h}_{0:T-1} = \mathcal{E}(x_{0:T-1}).
        \end{equation}
        As shown in Figure~\ref{fig:contextLM} (bottom), the token sequence is divided into fixed-length chunks of size $w$. These token embeddings serve two purposes: (i) providing fine-grained token representations for NTP, and (ii) supplying the foundation for constructing higher-level context embeddings. 
        
        \subsubsection{Context Predictor $\mathcal{P}$} 
        The Context Predictor $\mathcal{P}$ operates on a sequence of context embeddings derived from token representations.
        A mapping function $f(\cdot)$ aggregates token-level hidden states within each chunk:
        \begin{equation}
            \mathbf{c}_{0:K-1} = f(\mathbf{h}_{0:T-1}), \quad K=\lfloor T/w \rfloor+1.
        \end{equation}
        The Context Predictor then performs autoregressive modeling over this context sequence:
        \begin{equation}
            \mathbf{\hat{c}}_{1:K} = \mathcal{P}(\mathbf{c}_{0:K-1}),
        \end{equation}
        where each $\hat{c}_k$ depends only on previously inferred contexts. 
        
        To handle boundary cases where the first chunk has no preceding context available, we introduce an initialization embedding ${c}_{init}$, which is prepended to the context sequence and serves as a prior context embedding before the first prediction:
        \begin{equation}
            \mathbf{\hat{c}} = ({c}_{init}, \hat{c}_1, \ldots, \hat{c}_{K}).
        \end{equation}
        When fed ${c}_{init}$ into the Token Decoder, ${c}_{init}$ serves as the null hypothesis for the context information in the first chunk of tokens.
        
        Notably, context prediction occurs only at chunk boundaries (see Algorithm 1), leading to substantially fewer prediction steps than token-level modeling.
        
        \subsubsection{Token Decoder $\mathcal{D}$}
        The Token Decoder $\mathcal{D}$ performs standard NTP conditioning on both token embeddings and predicting context embeddings.
        To align granularities, each predicted context $\hat{c}_k$ is broadcast to all tokens within its corresponding chunk:
        \begin{equation}
            \hat{\mathbf{c}}^b = (\underbrace{{c}_{init}, \ldots, {c}_{init}}_{w-1\text{ times}}, \underbrace{\hat{c}_1, \ldots, \hat{c}_1}_{w\text{ times}}, \ldots, \underbrace{\hat{c}_{K-1}, \ldots, \hat{c}_{K-1}}_{T+1-(K-1)w\text{ times}}).
        \end{equation}
        The decoder then combines multi-level information via element-wise addition $h\oplus\hat{c}^b$. 
        The decoder then proceeds with causal attention and softmax-based token prediction.

    All components employ causal attention to strictly enforce autoregressive constraints and prevent information leakage. Temporal causality is guaranteed through consistent one-step lookahead logic at each hierarchical level, as illustrated in Figure~\ref{fig:contextLM}.
    At the token level, the Token Decoder follows the standard \textbf{one-token shift}: when predicting token $x_t$, the hidden state $h_t$ is conditioned on previously generated tokens $x_{<t}$. At the context level, tokens are partitioned into non-overlapping chunks, with each chunk $k$ summarized into a context embedding $c_k$. The Context Predictor applies a \textbf{one-chunk shift}, predicting $\hat{c}_k$ based solely on past context embedding $c_{<k}$. By aligning these two autoregressive processes across token-level and context-level modeling, ContextLM maintains full compatibility with standard NTP while strictly preventing information leakage.
    \subsection{Model Training}
    ContextLM retains the standard token-level cross-entropy loss $\mathcal{L}_{CE}$, ensuring compatibility with mainstream token-level models. In addition, unlike standard NTP where logits $z_t = \mathcal{D}(h_t)$ depend solely on local token representations, ContextLM conditions generation on predicted context embeddings: $z_t' = \mathcal{D}(h_t, \hat{c}_k)$. 
    This modification fundamentally reshapes the gradient flow during backpropagation, providing richer multi-level supervision while preserving the original training objective. Specifically:

    \begin{itemize}[leftmargin=*]

        \item \textbf{Context-level Supervision for Context Predictor.} Each predicted context embedding $\hat{c}_k$ conditions all tokens within its corresponding chunk set $\mathcal{J}_k$. Consequently, it receives an aggregated error signal summed across these token positions:
        \begin{equation}
            \frac{\partial \mathcal{L}_{CE}}{\partial \hat c_k} = \sum_{j \in \mathcal{J}_k} \frac{\partial \mathcal{L}_{CE}}{\partial z_j'} \frac{\partial z_j'}{\partial \hat c_k}.
        \end{equation}
        This aggregated supervision encourages the Context Predictor to capture high-level semantics relevant to the entire chunk rather than adapting to individual local patterns.

        \item \textbf{Multi-level Supervision for Token Encoder.} The Token Encoder parameters receive both token-level and context-level gradient signals. The gradient for $h_t$ is composed of the direct local signal and the indirect context signal:
        \begin{equation}
            \frac{\partial \mathcal{L}_{CE}}{\partial h_t} = 
            \underbrace{\frac{\partial \mathcal{L}_{CE}}{\partial z_t'} \frac{\partial z_t'}{\partial h_t}}_{\text{token-level signal}} 
            + 
            \underbrace{\Bigg(\sum_{j \in \mathcal{J}_k} \frac{\partial \mathcal{L}_{CE}}{\partial z_j'} \frac{\partial z_j'}{\partial \hat c_k}\Bigg) \frac{\partial \hat c_k}{\partial h_t}}_{\text{context-level signal}}.
        \end{equation}
        This ensures that the encoder representations are optimized not only for immediate NTP but also for constructing robust context representations that facilitate multi-step future generation. 
        \footnote{
        Even when the chunk size is set to $w=1$, ContextLM does not equal to standard NTP. 
        In this case, each token corresponds to a single context chunk.
        At step $t$, the model generates $x_{t+1}$ by conditioning on the token representation $h_t$ and a predictive context $\hat c_{t+1}$ (inferred from ${c_{\le t}}$), resulting in a hierarchical two-step prediction process.
        }

    \end{itemize}

\section{Experiments}

    We conduct extensive experiments to comprehensively assess the effectiveness, scalability, and generality of \MethodName\hspace{1pt} across different backbones, dataset scales, and evaluation settings. The experiments cover four main aspects: (i) scaling law behavior on both GPT-2 and Pythia families (Sec.~\ref{sec: Scaling Experiments}); (ii) comprehensive evaluation on diverse downstream tasks (Sec.~\ref{sec: Downstream Task Evaluation}); (iii) instruction-following performance after fine-tuning (Sec.~\ref{sec: Instruction-following Ability Evaluation}); and (iv) ablation studies and analysis (Sec.~\ref{sec: Ablation and Analysis}).

    \subsection{Experimental Setup}
    \paragraph{Models and Datasets} We evaluate \MethodName\hspace{1pt} on two widely used Transformer families to demonstrate its architectural compatibility and effectiveness. (i) \textbf{GPT-2}: models are pretrained from scratch on the OpenWebText~\citep{Gokaslan2019OpenWeb} with a maximum sequence length of 1024. Additional implementation details on GPT-2 are provided in Appendix~\ref{appendix:training_hyper}. (ii) \textbf{Pythia}: to examine large-scale scalability, we train ContextLM-Pythia from scratch on the entire Pile dataset~\citep{gao2020pile} (300B tokens) with a maximum sequence length of 2048, following the original Pythia tokenizer and training configuration (optimizer settings, learning rate schedule, batch size, and context length).  

    \paragraph{ContextLM Settings} In our practice, we use the mean pooling function as the mapping function $f(\cdot)$ in Sec.~\ref{sec: architecture}. We use the first token embedding $h_0$ as ${c}_{init}$. 
    For Pythia models that use PoSE as the position embedding technique, we use the position of the first token in each chunk as the context chunk position in context layers. Unless specifically noted, we set the context chunk size to $w=4$ and use a two-layer Context Predictor. 
    
    \subsection{Scaling Experiments}
    \label{sec: Scaling Experiments}
    \paragraph{Scaling Law on GPT-2}
    \label{sec: Scaling Law on GPT2}
    We begin by evaluating the scaling behavior of ContextLM on the GPT-2 family, referred to ContextLM-GPT2, trained on OpenWebText under matched compute budgets of the baseline GPT-2. 
    Figure~\ref{fig:scaling_law} reports the perplexity as a function of model parameters, training tokens, and training FLOPs. Across all three scaling dimensions, our model consistently outperforms the GPT-2 baseline. 
    Moreover, we compare ContextLM against the Parameter-Matched (PM) baselines, constructed by adding two standard Transformer layers, in Table~\ref{tab:pm-model_performance}.
    These results demonstrate the favorable scaling of our method.
    \begin{table*}[ht]
        \centering
        \caption{Performance comparison ContextLM with baseline on GPT-2 backbone across different model scales. See Appendix~\ref{appendix:downsream_task} for full task details.}
        \renewcommand{\arraystretch}{1.3}
        \scalebox{1}{
        \begin{tabular}{llll}
        \toprule
        \multirow{2}{*}{Model} & \multirow{2}{*}{Avg PPL$\downarrow$} & \multicolumn{2}{c}{Avg Acc $\uparrow$ } \\
        \cmidrule(lr){3-4}
        & &{0-shot} & {5-shot}  \\ \midrule
        GPT2-Base-124M & 110.92 & 37.8 & 36.6 \\
        GPT2-Base-PM & 107.77 & 38.0 & 36.7 \\
        \rowcolor[HTML]{F2F3F5}
        \textbf{ContextLM-Base} & \textbf{87.19}\textsubscript{\textcolor{green!60!black}{$\downarrow$23.73}} & \textbf{39.1}\textsubscript{\textcolor{green!60!black}{$\uparrow$1.3}} & \textbf{37.6}\textsubscript{\textcolor{green!60!black}{$\uparrow$1.0}} \\ \midrule
        GPT2-Medium-355M & 55.46 & 40.5 & 38.8 \\
        GPT2-Medium-PM & 53.18 & 41.0 & 39.4 \\
        \rowcolor[HTML]{F2F3F5}
        \textbf{ContextLM-Medium} & \textbf{43.06}\textsubscript{\textcolor{green!60!black}{$\downarrow$12.40}} & \textbf{41.9}\textsubscript{\textcolor{green!60!black}{$\uparrow$1.4}} & \textbf{40.2}\textsubscript{\textcolor{green!60!black}{$\uparrow$1.4}} \\ \midrule
        GPT2-Large-774M & 38.27 & 41.4 & 41.0\\
        GPT2-Large-PM & 36.28 & 42.4 & 41.3 \\
        \rowcolor[HTML]{F2F3F5}
        \textbf{ContextLM-Large} & \textbf{31.80}\textsubscript{\textcolor{green!60!black}{$\downarrow$6.47}} & \textbf{43.9}\textsubscript{\textcolor{green!60!black}{$\uparrow$2.5}} & \textbf{43.1}\textsubscript{\textcolor{green!60!black}{$\uparrow$2.1}} \\ \midrule
        GPT2-XL-1.5B & 31.54 & 43.3 & 42.1\\
        GPT2-XL-PM & 31.57 & 43.2 & 42.4 \\
        \rowcolor[HTML]{F2F3F5}
        \textbf{ContextLM-XL} & \textbf{25.93}\textsubscript{\textcolor{green!60!black}{$\downarrow$5.61}} & \textbf{45.0}\textsubscript{\textcolor{green!60!black}{$\uparrow$1.7}} & \textbf{44.0}\textsubscript{\textcolor{green!60!black}{$\uparrow$1.9}} 
        \\ \bottomrule
        \end{tabular}}
        \label{tab:pm-model_performance}
    \end{table*}
 
    

    \paragraph{Scaling Law on Pythia}
    \label{sec: Scaling Law on Pythia}
    To examine the general applicability and effectiveness of ContextLM, we further evaluate it on the Pythia family ($70$M–$1.4$B parameters) to assess scalability across larger data (300B tokens). As shown in Table~\ref{contextlm_results_ppl}, ContextLM-Pythia consistently achieves lower perplexity across the Pile, Wikitext~\citep{merity2016pointer}, and Lambada~\citep{paperno2016lambada}. 
    The relative improvement is most pronounced at smaller scales and remains consistent as model size increases, confirming that the benefits of context-level supervision persist throughout scaling. These results demonstrate that \MethodName\ maintains both scalability and computational efficiency on large-scale datasets.
       \begin{table*}[ht]
        \centering
        \renewcommand{\arraystretch}{1.3}
        \caption{Perplexity Comparison of Pythia and ContextLM-Pythia across four benchmark datasets. For brevity, we denote ContextLM-Pythia as ContextLM. ``Avg PPL'' reports the average perplexity per model, where lower values indicate better performance.}
        \label{tab:results}
        \resizebox{1\textwidth}{!}{
        \begin{tabular}{lllll|l}
        \toprule
        Model & Pile & Wikitext & Lambada OpenAI & Lambada Standard & Avg PPL $\downarrow$\\
        \midrule
        Pythia-70M    & 18.27 & 57.01 & 142.01 & 973.59 & 297.72 \\
        \rowcolor[HTML]{F2F3F5}
        \textbf{ContextLM-70M} & \textbf{14.96\textcolor{green!60!black}{\scriptsize{${\downarrow3.31}$}}} & \textbf{43.64\textcolor{green!60!black}{\scriptsize{${\downarrow13.37}$}}} & \textbf{71.45\textcolor{green!60!black}{\scriptsize{${\downarrow70.56}$}}} & \textbf{440.77\textcolor{green!60!black}{\scriptsize{${\downarrow532.82}$}}} & \textbf{142.71\textcolor{green!60!black}{\scriptsize{${\downarrow155.01}$}}} \\
        \midrule
        Pythia-160M & 12.56 & 33.44 & 38.20 & 187.28 & 67.87 \\
        \rowcolor[HTML]{F2F3F5}
        \textbf{ContextLM-160M} & \textbf{11.14\textcolor{green!60!black}{\scriptsize{${\downarrow1.42}$}}} & \textbf{28.18\textcolor{green!60!black}{\scriptsize{${\downarrow5.26}$}}} & \textbf{25.97\textcolor{green!60!black}{\scriptsize{${\downarrow12.23}$}}} & \textbf{107.05\textcolor{green!60!black}{\scriptsize{${\downarrow80.23}$}}} & \textbf{43.09\textcolor{green!60!black}{\scriptsize{${\downarrow24.78}$}}} \\
        \midrule
        Pythia-410M & 8.88 & 20.11 & 10.85 & 31.53 & 17.84 \\
        \rowcolor[HTML]{F2F3F5}
        \textbf{ContextLM-410M} & \textbf{8.67\textcolor{green!60!black}{\scriptsize{${\downarrow0.21}$}}} & \textbf{19.50\textcolor{green!60!black}{\scriptsize{${\downarrow0.61}$}}} & \textbf{10.15\textcolor{green!60!black}{\scriptsize{${\downarrow0.70}$}}} & \textbf{24.95\textcolor{green!60!black}{\scriptsize{${\downarrow6.58}$}}} & \textbf{15.82\textcolor{green!60!black}{\scriptsize{${\downarrow2.02}$}}}\\
        \midrule
        Pythia-1B & 7.82 & 16.45 & 7.92 & 17.44 & 12.41 \\
        \rowcolor[HTML]{F2F3F5}
        \textbf{ContextLM-1B} & \textbf{7.66\textcolor{green!60!black}{\scriptsize{${\downarrow0.16}$}}} & \textbf{16.09\textcolor{green!60!black}{\scriptsize{${\downarrow0.36}$}}} & \textbf{7.38\textcolor{green!60!black}{\scriptsize{${\downarrow0.54}$}}} & \textbf{15.75\textcolor{green!60!black}{\scriptsize{${\downarrow1.69}$}}} & \textbf{11.72\textcolor{green!60!black}{\scriptsize{${\downarrow0.69}$}}} \\
        
        \midrule
        Pythia-1.4B & 7.26 & 14.72 & 6.09 & 10.87 & 9.74 \\
        \rowcolor[HTML]{F2F3F5}  
        \textbf{ContextLM-1.4B} & \textbf{7.16\textcolor{green!60!black}{\scriptsize{${\downarrow0.10}$}}} & \textbf{14.61\textcolor{green!60!black}{\scriptsize{${\downarrow0.11}$}}} & \textbf{6.06\textcolor{green!60!black}{\scriptsize{${\downarrow0.03}$}}} & \textbf{10.30\textcolor{green!60!black}{\scriptsize{${\downarrow0.57}$}}} & \textbf{9.54\textcolor{green!60!black}{\scriptsize{${\downarrow0.20}$}}}\\
        \bottomrule
        \end{tabular}
        }
        \label{contextlm_results_ppl}
    \end{table*}

    \subsection{Downstream Task}
    \label{sec: Downstream Task Evaluation}
    To evaluate generalization beyond perplexity, we evaluate the zero-shot and five-shot performance using the \texttt{lm-evaluation-harness}~\footnote{https://github.com/EleutherAI/lm-evaluation-harness} across 9 widely-used benchmarks, which we categorize into three core capabilities: linguistic understanding (Lambada OpenAI/Standard~\citep{paperno2016lambada}, WinoGrande~\citep{Sakaguchi2019AnAW}), commonsense reasoning (ARC-Easy/Challenge~\citep{Clark2018ThinkYH}, PIQA~\citep{Bisk2019PIQARA}, HellaSwag~\citep{Zellers2019HellaSwagCA}, SciQ~\citep{SciQ}), and complex reasoning (RACE~\citep{race}), following the evaluation established in~\citep{gu2024mamba,Zeng2025PretrainingLM}. 

    As shown in Table~\ref{tab:contextlm_results_acc}, ContextLM-Pythia consistently outperforms the Pythia baseline across all model sizes in both evaluation settings. 
    Improvements are observed across all task categories, with larger relative gains at smaller scales and more stable absolute gains at larger scales. Notably, the advantages are more pronounced in the five-shot setting, suggesting that context-level supervision yields representations that transfer more effectively under limited in-context supervision. Additional downstream results on ContextLM-GPT2, GPT-2, and PM-GPT2 baselines are reported in Appendix~\ref{appendix:downsream_task}. Across both GPT-2 and Pythia backbones, ContextLM consistently improves downstream performance over baselines, demonstrating that context-level supervision better utilizes available parameters and generalizes across diverse tasks.

       \begin{table*}[ht]
    \centering
    \caption{Zero-shot and five-shot evaluation results across 9 downstream benchmarks. 
    ``Avg Acc'' denotes the mean accuracy per model across each evaluation condition, with higher values indicating better performance.}
    \label{tab:accuracy_results}
    \renewcommand{\arraystretch}{1.3}
    \resizebox{1\textwidth}{!}{
    \begin{tabular}{lccccccccc|c}
    \toprule
    Model & \makecell{Lambada \\ OpenAI} & \makecell{ARC-\\E} & \makecell{Lambada \\ Standard} & \makecell{ARC-\\C} & \makecell{Wino\\Grande} & \makecell{PIQA} & \makecell{Hella-\\Swag} & \makecell{SciQ} & \makecell{RACE} & \makecell{Avg \\ Acc }$\uparrow$ \\
    \midrule
    \midrule
    \rowcolor[HTML]{E3E4E5}
    \multicolumn{11}{c}{\textit{{\textbf{Zero-shot}}}}\\
    \midrule
    Pythia-70M & 18.3 & 36.9 & 13.4& 18.5& 52.1 & 60.0& 26.6 & 60.5& 24.9 & 34.6\\
    \rowcolor[HTML]{F2F3F5}
    \textbf{ContextLM-70M} &  28.0 & 40.7 & 17.2 & 18.6 & 52.3 & 60.5 & 27.5 & 72.1 & 27.0 & \textbf{38.2 / \textcolor{green!60!black}{\textbf{${\uparrow3.6}$}}}\\
    \midrule
    Pythia-160M & 32.7 & 43.8 & 21.5 & 19.5 & 53.4 & 61.5 & 28.5 & 74.3 & 27.9 & 40.3 \\
    \rowcolor[HTML]{F2F3F5}
    \textbf{ContextLM-160M} & 37.2 & 45.0 & 25.6 & 19.5 & 52.0 & 62.9 & 29.2 & 77.6 & 28.7 & \textbf{42.0 / \textcolor{green!60!black}{\textbf{${\uparrow1.7}$}}} \\
    \midrule
    Pythia-410M & 51.6 & 52.2 & 36.4 & 21.3 & 53.9 & 66.8 & 33.8 & 81.2 & 30.7 & 47.5 \\
    \rowcolor[HTML]{F2F3F5}
    \textbf{ContextLM-410M} & 52.2 & 51.7 & 39.0 & 22.7 & 52.3 & 67.4 & 34.3 & 83.0 & 30.8 & \textbf{48.2 / \textcolor{green!60!black}{\textbf{${\uparrow0.7}$}}} \\
    \midrule
    Pythia-1B & 55.9 & 56.8 & 42.0 & 24.2 & 52.5 & 70.5 & 37.7 & 83.3 & 32.7 & 50.6 \\
    \rowcolor[HTML]{F2F3F5}
    \textbf{ContextLM-1B} & 57.8  & 55.3  & 43.7  & 25.9  & 55.0  & 70.1  & 37.6  & 86.1  & 32.1  & \textbf{51.5 / \textcolor{green!60!black}{\textbf{${\uparrow0.9}$}}}\\
    \midrule
    Pythia-1.4B & 61.6 & 60.4 & 49.7 & 25.9 & 57.5 & 70.8 & 40.4 & 86.4 & 34.1 & 54.1 \\
    \rowcolor[HTML]{F2F3F5}
    \textbf{ContextLM-1.4B} & 61.6 & 58.9 & 51.4 & 27.2 & 55.7 & 71.2 & 40.6 & 87.9 & 34.9 & \textbf{54.4 / \textcolor{green!60!black}{\textbf{${\uparrow0.3}$}}}  \\
    \midrule
    \midrule
    \rowcolor[HTML]{E3E4E5}
    \multicolumn{11}{c}{\textit{{\textbf{Five-shot}}}}\\
    \midrule
    Pythia-70M     & 11.9 & 36.7 & 9.2  & 17.1 & 50.5 & 58.7 & 26.7 & 57.8 & 25.1 & 32.6 \\
    \rowcolor[HTML]{F2F3F5}
    \textbf{ContextLM-70M}  & 19.2 & 41.4 & 14.2 & 18.5 & 51.1 & 60.8 & 27.7 & 71.5 & 25.7 & \textbf{36.7 / \textcolor{green!60!black}{\textbf{${\uparrow4.1}$}}}\\
    \midrule
    Pythia-160M    & 24.9 & 44.7 & 19.0 & 18.4 & 50.4 & 63.5 & 28.6 & 76.4 & 27.8 & 39.3 \\
    \rowcolor[HTML]{F2F3F5}
    \textbf{ContextLM-160M} & 29.1 & 45.7 & 23.1 & 19.8 & 51.9 & 63.1 & 29.5 & 81.7 & 28.2 &\textbf{41.3 / \textcolor{green!60!black}{\textbf{${\uparrow2.0}$}}}\\
    \midrule
    Pythia-410M    & 43.9 & 54.7 & 32.8 & 22.3 & 53.4 & 68.0 & 33.8 & 88.9 & 30.4 & 47.6 \\
    \rowcolor[HTML]{F2F3F5}
    \textbf{ContextLM-410M} & 44.6  & 54.8  & 34.9  & 23.0  & 52.7  & 68.0  & 34.3  & 89.5  & 30.9  & \textbf{48.1 / \textcolor{green!60!black}{\textbf{${\uparrow0.5}$}}} \\
    \midrule
    Pythia-1B   & 48.3 & 58.6 & 35.8 & 25.4 & 52.8 & 71.3 & 37.7 & 91.6 & 31.7 & 50.4 \\
    \rowcolor[HTML]{F2F3F5}
    \textbf{ContextLM-1B } & 49.9  & 60.2  & 39.5  & 24.2  & 54.9  & 70.5  & 38.1  & 91.5  & 32.6  &\textbf{51.3 / \textcolor{green!60!black}{\textbf{${\uparrow0.9}$}}} \\
    \midrule
    Pythia-1.4B    & 54.5 & 63.1 & 44.5 & 28.8 & 57.1 & 71.0 & 40.5 & 92.4 & 34.6 & 54.1 \\
    \rowcolor[HTML]{F2F3F5}
    \textbf{ContextLM-1.4B} & 55.7 & 62.3 & 46.7 & 28.5 & 56.8 & 72.4 & 41.1 & 93.3 & 35.0 & \textbf{54.6 / \textcolor{green!60!black}{\textbf{${\uparrow0.5}$}}}  \\
    \bottomrule
    \end{tabular}
    }
    \label{tab:contextlm_results_acc}
    \end{table*}

    \subsection{Instruction-following Ability}
    \label{sec: Instruction-following Ability Evaluation}
\begin{figure*}[htb]
      \centering
      \scalebox{1}{
      \begin{subfigure}{0.42\textwidth}
        \includegraphics[width=\linewidth]{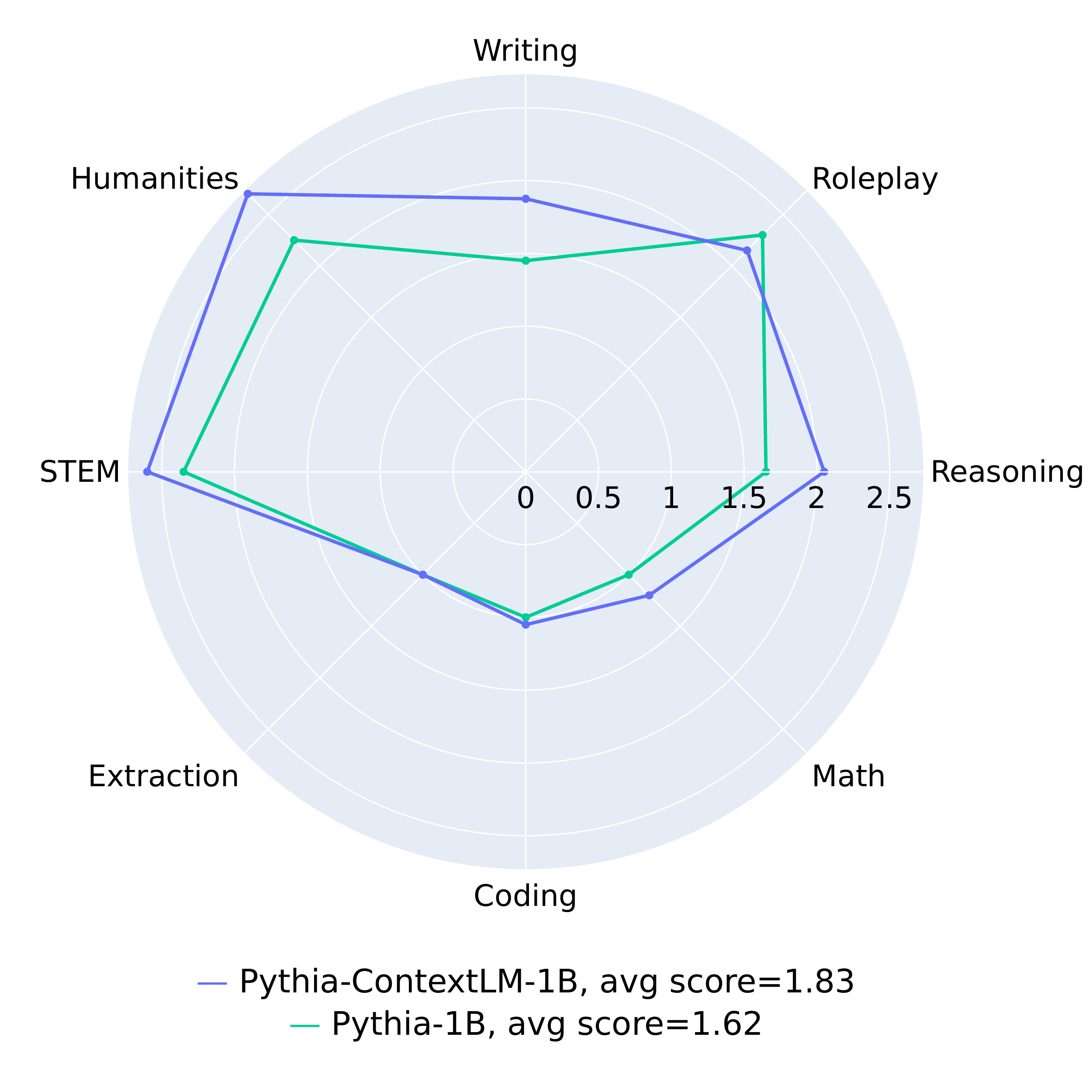}
      \end{subfigure}
      \hspace{30pt}
      \begin{subfigure}{0.42\textwidth}
          \includegraphics[width=\linewidth]{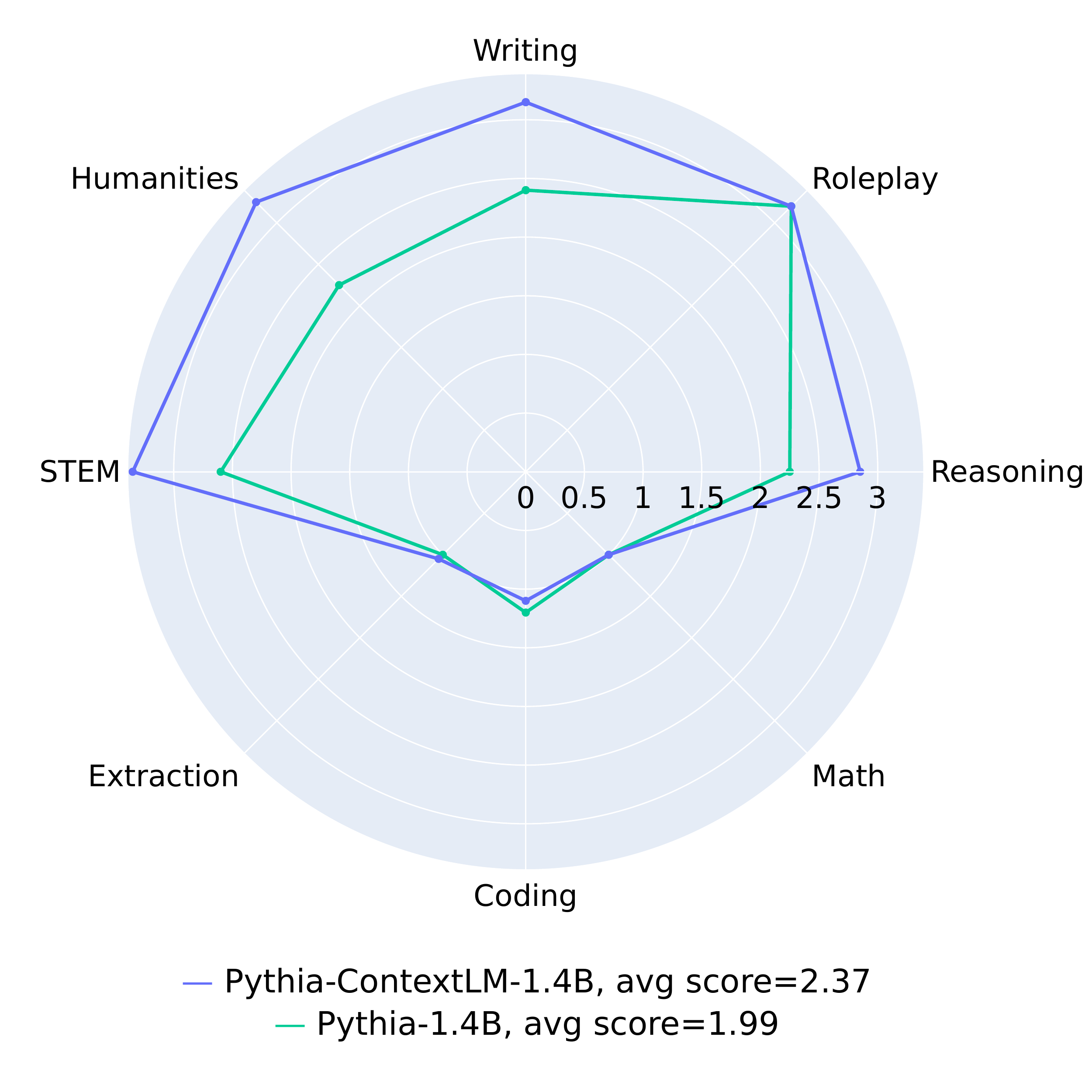}
      \end{subfigure}}
      \caption{Instruction-following evaluation on MT-Bench across multiple subtasks. 
      }
    \label{fig:instruction_following}
    \end{figure*}

    To assess the instruction-following ability of our model, we further fine-tuned ContextLM-Pythia and official Pythia on the Alpaca~\citep{taori2023stanford} using the same settings~\citep{taori2023stanford}, and evaluated using MT-Bench~\citep{zheng2023judging} and AlpacaEval 2.0~\citep{dubois2024length}. As shown in Figure~\ref{fig:instruction_following}, ContextLM-Pythia consistently surpasses Pythia across multiple capability subtasks. At the $1$B scale, the average score improves from 1.62 to 1.83, while the $1.4$B model shows more substantial gains, increasing from 1.99 to 2.37. 
    On the more modern AlpacaEval 2.0 benchmark (Table~\ref{tab:alpaca_eval}), ContextLM achieves a 54.48\% win rate over the Pythia-1.4B baseline. These results suggest that context-level supervision not only enhances pretraining efficiency but also facilitates stronger instruction-following performance after fine-tuning.

     \begin{table*}[ht]
    \centering
    \caption{{AlpacaEval 2.0 Win Rate Comparison.} Evaluated using Pythia-1.4B-SFT as the reference model.}
    \label{tab:alpaca_eval}
    \scalebox{1}{
    \begin{tabular}{lcc}
    \toprule
    {Model} & {Win Rate (\%)} & {LC Win Rate (\%)} \\
    \midrule
    Pythia-1.4B & 45.52 & 45.99 \\
    \rowcolor[HTML]{F2F3F5}
    \textbf{ContextLM-Pythia-1.4B} & \textbf{54.48} & \textbf{54.01} \\
    \bottomrule
    \end{tabular}
    }
    \end{table*}

    \section{Analysis}
    \label{sec: Ablation and Analysis}
    In this section, we conduct a comprehensive analysis 
    to understand how key choices in ContextLM influence model behavior and empirical performance.

    \subsection{Ablation Study}
    \label{ablation: chunk size and Decoder Depth}
\begin{figure*}[htb]
        \centering
        \begin{subfigure}{0.245\textwidth}
            \includegraphics[width=\linewidth]{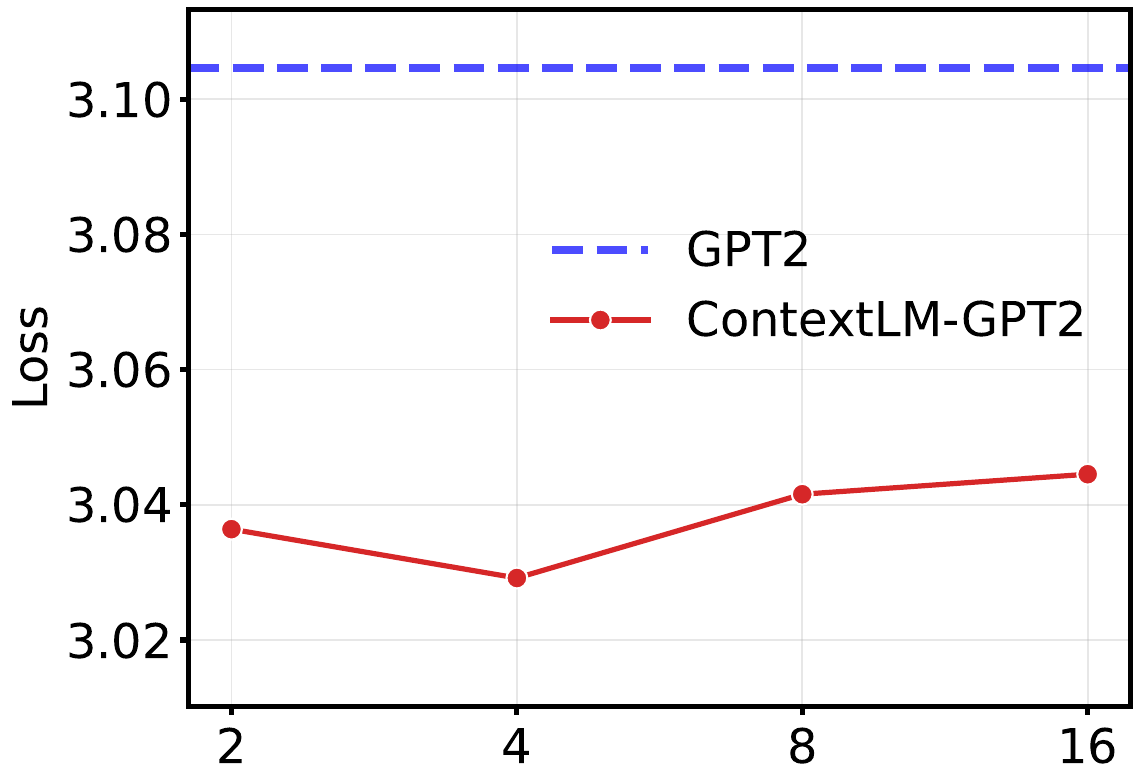}
            \subcaption{Chunk Size}
        \end{subfigure}
        \begin{subfigure}{0.245\textwidth}
            \includegraphics[width=\linewidth]{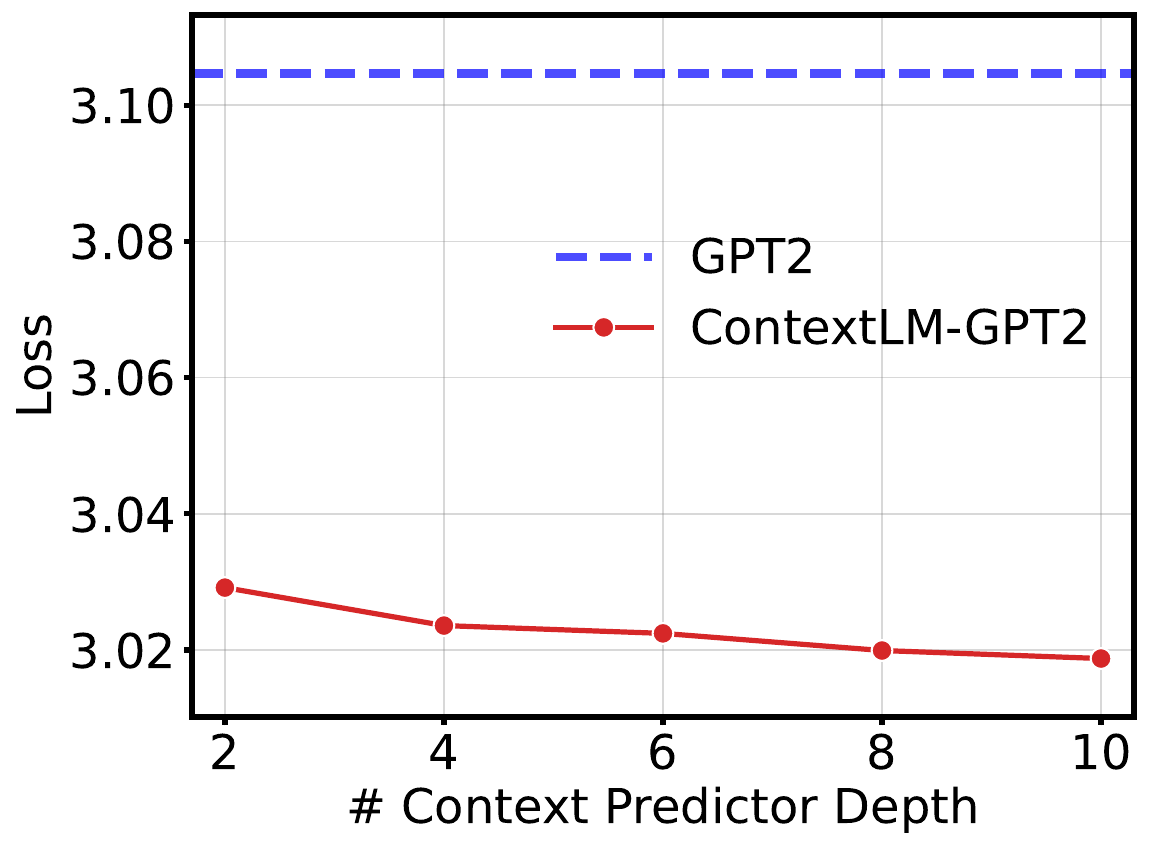}
            \subcaption{Predictor Depth}
        \end{subfigure}
        \begin{subfigure}{0.245\textwidth}
            \includegraphics[width=\linewidth]{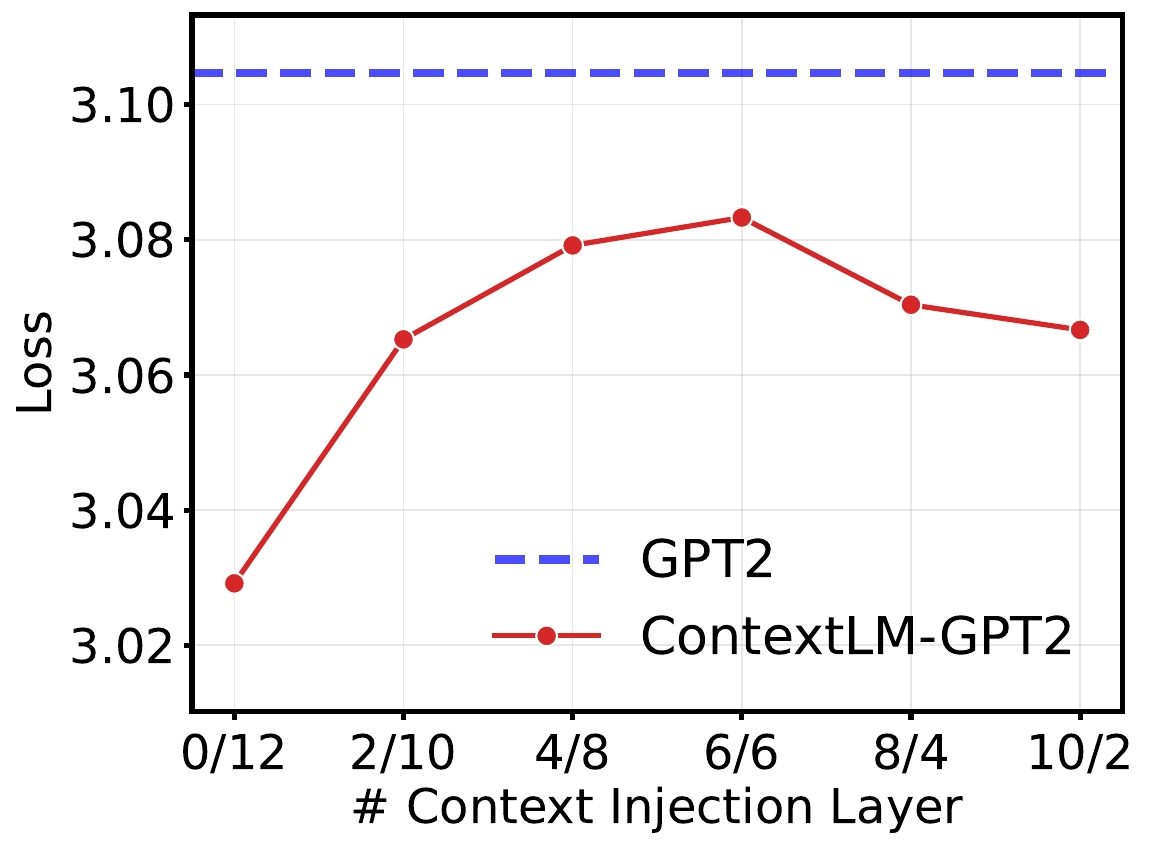}
            \subcaption{Injection Layer}
        \end{subfigure}
        \begin{subfigure}{0.245\textwidth}
            \includegraphics[width=\linewidth]
            {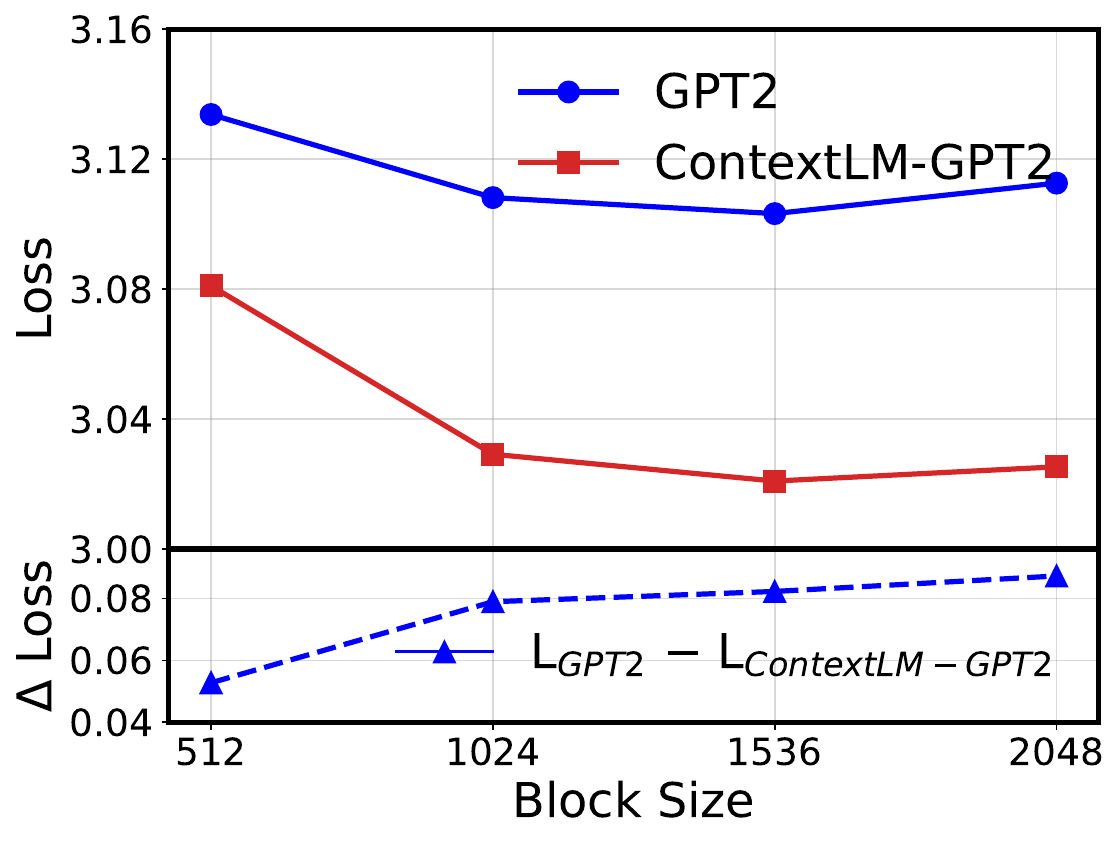} 
            \subcaption{Block Size}
            
        \end{subfigure}
        \caption{
        Validation loss on OpenWebText for baseline GPT2-Base and ContextLM-GPT2-Base across chunk sizes, Context Predictor depths, context injection layer, and block size.
        }
    \label{fig:ablation}
    \end{figure*}

    \paragraph{Chunk Size} We first study the effect of chunk size $w$ by varying it across $\{2, 4, 8, 16\}$. To enable a fair comparison under comparable computational budgets, we adjust the depth of the Context Predictor with respect to $w$, as illustrated in Figure~\ref{fig:ablation} (a). 
    Specifically, as increasing $w$ reduces the effective sequence length processed by the predictor, we increase its depth to keep the overall FLOPs and memory approximately constant.
    Our results indicate that $w=4$ offers the best trade-off, so we adopt $w=4$ as the default setting.

    \paragraph{Context Predictor Depth} We next analyze the effect of the Context Predictor depth (Figure~\ref{fig:ablation} (b)). 
    Increasing the predictor depth beyond two layers results in only marginal gains, indicating that a 2-layer predictor is sufficient to capture the dominant semantic transitions for effective context-level modeling; we therefore adopt a 2-layer Context Predictor as the default configuration.
    
    \paragraph{Context Injection Layer} 
    We investigate where to inject the Context Predictor by varying the split layer between the Token Encoder and Token Decoder.
    As shown in Figure~\ref{fig:ablation} (c), the 0/12 setting provides the best performance.
    This configuration allows more decoder layers to process and integrate the contextual signal, leading to deeper semantic integration. Furthermore, since shallower layers primarily encode local and low-level features, the predictor must therefore infer high-level semantics from limited contextual cues, making the learning task more challenging.
    
    \paragraph{Block Size}
    We further examine the effect of block size ranging from $512$ to $2048$ tokens (Figure~\ref{fig:ablation} (d)). Since ContextLM operates on chunked representations, the effective sequence length is reduced by a factor of $w$, allowing more efficient context modeling of long-range context.
    As block size increases, the $\Delta$Loss between ContextLM-GPT2 and GPT-2 widens, indicating that context-level supervision provides increasing benefits as the context length grows.
    We extend our evaluation to long-context reasoning using the Pythia-1.4B backbone on LongBench~\citep{bai2024longbench}. To handle input sequences exceeding the model's pre-training limit of $2048$ tokens, we directly apply dynamic NTK-aware RoPE interpolation. As shown in Table~\ref{tab:longbench_summary}, ContextLM consistently outperforms the baseline across all subtasks, achieving a +3.11 improvement in overall average accuracy. 
    Similar trends are observed on LongBench, supporting that the improvements extend beyond validation loss to evaluations involving longer input contexts.

    \begin{table*}[h]
    \centering
    \caption{{LongBench Performance Summary (0-4k Subtasks).} ContextLM demonstrates superior long-context reasoning capabilities, consistently outperforming the Pythia-1.4B baseline across diverse task categories.}
    \label{tab:longbench_summary}
    \renewcommand{\arraystretch}{1.3}
    \begin{tabular}{lcccccc}
    \toprule
    {Model} & {Single-Doc} & {Multi-Doc} & {Summary} & {Synthetic} & \makecell{Code \\ \& Class.} & \makecell{Overall \\ Avg }$\uparrow$ \\
    \midrule
    Pythia-1.4B & 13.17 & 4.49 & 13.97 & 2.32 & 33.94 & 20.21 \\
    \rowcolor[HTML]{F2F3F5}
    \textbf{ContextLM-Pythia-1.4B} & \textbf{13.41} & \textbf{4.92} & \textbf{20.19} & \textbf{4.39} & \textbf{40.40} & \textbf{23.32} \\
    \bottomrule
    \end{tabular}
    \end{table*}

    \subsection{Comparison with Multi-Token Prediction}
    To evaluate the benefit of context-level modeling over extending the token prediction horizon, we compare ContextLM with standard multi-token prediction (MTP). Following official implementation settings~\citep{gloeckle2024better}, we configure MTP to predict $4$ future tokens, matching the ContextLM chunk size $w=4$. Experimental results on the GPT2-XL scale (Table~\ref{tab:mtp_comparison}) show that ContextLM significantly outperforms both NTP and MTP in terms of perplexity and downstream task performance. 
    These results suggest that predicting higher-level contextual representations provides a more informative supervision signal than extending the prediction horizon through multi-token prediction.
    \begin{table*}[ht]
    \centering
    \caption{
    Comparison against Multi-Token Prediction (MTP) at the GPT2-XL scale trained on OpenWebText.
    }
    \label{tab:mtp_comparison}
    \setlength{\tabcolsep}{1.3pt}  
    \resizebox{\textwidth}{!}{
    \begin{tabular}{l|c|ccccccccc|c}
    \toprule
    Model &  PPL $\downarrow$ & \makecell{Lambada \\ OpenAI} & \makecell{ARC-\\E} & \makecell{Lambada \\ Standard} & \makecell{ARC-\\C} & \makecell{Wino\\Grande} & \makecell{PIQA} & \makecell{Hella-\\Swag} & \makecell{SciQ} & \makecell{RACE} & \makecell{Avg \\ Acc }$\uparrow$ \\
    \midrule
    GPT2-XL-NTP & 15.25 & 38.7 & 48.1 & 29.2 & 20.6 & 50.8 & 65.2 & 31.0 & 77.3 & 29.2 & 43.3 \\
    GPT2-XL-MTP & 16.72 & 36.7 & 49.1 & 25.4 & 19.6 & 49.3 & 64.1 & 30.3 & 77.4 & 30.5 & 42.5 \\
    \rowcolor[HTML]{F2F3F5}
    \textbf{ContextLM-XL} & \textbf{14.60 / \textcolor{green!60!black}{\textbf{${\downarrow 0.65}$}}} & \textbf{41.9} & \textbf{49.9} & \textbf{32.3} & \textbf{21.6} & \textbf{51.9} & \textbf{66.1} & \textbf{32.3} & \textbf{77.7} & \textbf{31.2} & \textbf{45.0 / \textcolor{green!60!black}{\textbf{${\uparrow 1.7}$}}} \\
    \bottomrule
    \end{tabular}
    }
    \end{table*}

    \subsection{Visualizing Attention Distribution}
    \label{sec: visualization}

\begin{figure*}[htb]
        \centering
        \includegraphics[width=\linewidth]{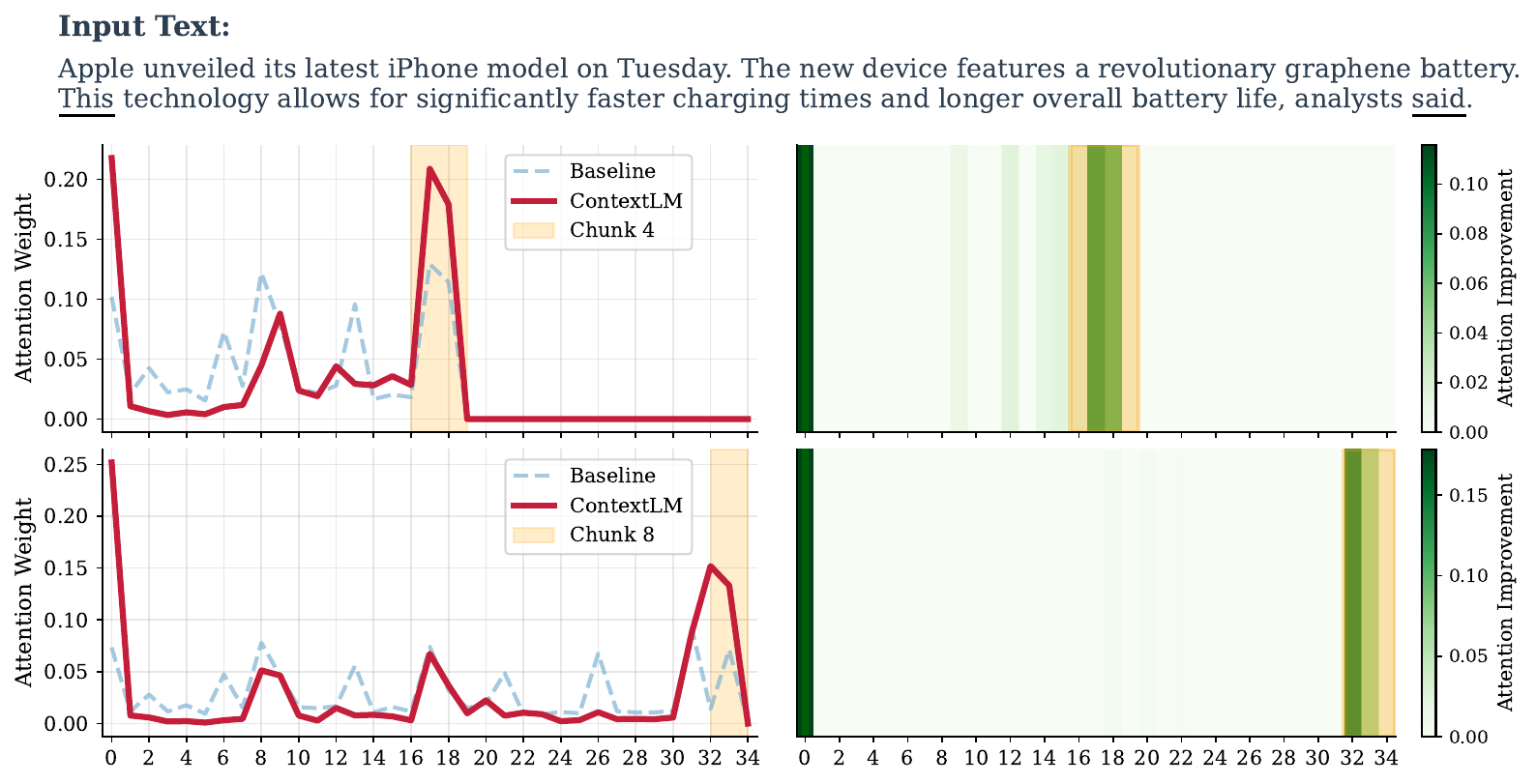}

        \caption{Attention weight analysis for the input text. ContextLM-GPT2-XL (red) shows a significant increase in attention over the baseline GPT2-XL (blue) towards the technical concept in Chunk 4 ("\textit{battery. \underline{This} technology}") and the contextual framing in Chunk 8 ("\textit{analysts \underline{said}.}"), indicating improved contextual understanding.}
        \label{fig:attention vis}

    \end{figure*}

    To qualitatively understand ContextLM’s behavior, we visualize a representative example in Figure~\ref{fig:attention vis}. 
    Compared to baseline, ContextLM allocates attention more selectively toward tokens that are salient for maintaining coherence across broader context.
    In particular, ContextLM places substantially higher attention on the anaphoric reference “\textit{this}” and the preceding technical concept “\textit{battery},” indicating increased reliance on chunk-level contextual information when predicting local tokens.
    A similar effect is observed for the reporting phrase “\textit{analysts said},” where ContextLM assigns moderately higher attention to the surrounding these tokens (approximately a 16\% increase).
    Overall, this pattern suggests that by incorporating predictive context embeddings, ContextLM enhances its ability to capture both token-level dependencies and high-level contextual relationships.
    

\section{Related Work}
    \subsection{Modeling the semantic level above tokens}
    Recent studies have explored multi-level architectures that extend beyond token-level next-token prediction (NTP) to explicitly model higher-level semantics.
    Block Transformer~\citep{ho2024block} introduces a global-to-local modeling strategy that accelerates inference while maintaining accuracy, implicitly learning sentence-level abstractions.
    LCM~\citep{lcmteam2024largeconceptmodelslanguage} employs diffusion models to predict conceptual semantics, but yielded limited performance gains and is constrained to a pre-defined semantic space.
    DLCM~\citep{qu2025dynamic} introduces latent representations within an adaptive semantic space, 
    while CALM~\citep{shao2025continuous} utilizes continuous abstract states to bridge discrete tokens and high-level contextual structures.
    In parallel, MTP~\citep{gloeckle2024better} and patch-based multi-token prediction~\citep{shao2024beyond} also move beyond strict token-by-token prediction, though their focus remains on optimizing the predictive objective rather than building hierarchical representations.
    ContextLM differs by making high-level context prediction an explicit component of the generative process—predicting and integrating latent context embeddings at chunk boundaries—while remaining fully compatible with the standard token-level autoregressive paradigm.

    \subsection{Modeling the semantic level below tokens}
    At a finer granularity, several approaches construct hierarchies by grouping characters or subword units into local patches.
    MegaByte~\citep{yu2023megabyte} adopts a multi-scale decoder that processes sequences at both byte and token levels, primarily targeting computational efficiency.
    BLT~\citep{pagnoni2024byte} learns patch boundaries using an entropy-based criterion, forming character-level groupings that improve modeling efficiency.
    H-Net~\citep{Hwang2025DynamicCF} dynamically adjusts patch sizes during training through an intermediate smoothing mechanism.
    These methods emphasize efficiency and scalability at lower granularity, whereas ContextLM targets higher-level latent representations through explicit context prediction—complementing rather than replacing token-level generation.
    


\section{Conclusion}

We introduce ContextLM, a hierarchical language modeling framework that learns predictive, context-level representations while remaining fully compatible with standard token-level evaluation metrics. By introducing a Context Predictor that autoregressively models higher-level context embeddings, our approach enables multi-token prediction implicitly and leverages aggregated error signals from future tokens to capture long-range dependencies. Experiments on the GPT-2 and Pythia families, up to $1.5$B parameters, show consistent improvements in perplexity and downstream performance, with gains persisting across model scales. Further analysis demonstrates that context-level supervision strengthens long-range coherence and attention allocation, highlighting next-context prediction as a scalable and efficient direction for advancing large language models.
\section*{Acknowledgment}
This work is sponsored by the National Natural Science Foundation of China (NSFC) grant (No. 62576211) and the National Key Research and Development Program of China (No. 2023ZD0121402). It is also the result of a collaborative project on novel language model architectures between Shanghai Jiao Tong University (SJTU) and the Shanghai Artificial Intelligence Laboratory. The computational resources required for pretraining the models were provided by the Shanghai AI Lab.
\bibliography{main}
\newpage
\section{Appendix}

    \subsection{Training Hyperparameters on GPT-2}
    \label{appendix:training_hyper}
    In our scaling law experiments, we adopt the configurations detailed in Table~\ref{tab:gpt2_hyper} for the GPT-2 model family. All models are trained on the OpenWebText with a maximum sequence length of 1024 tokens. Optimization is performed using AdamW ($\beta_1 = 0.9$, $\beta_2 = 0.95$) with gradient clipping at 1.0 and linear warmup over the first 1,000 steps. Learning rates are scaled according to model size following established practices.
        \begin{table}[ht]
            \centering
            \caption{Training hyperparameters for GPT-2 family models. 
            }
            \scalebox{1}{
            \begin{tabular}{lccccc}
            \toprule
            Model & $n_{\mathrm{head}}$ & $d_{\mathrm{model}}$ & {learning rate} & {batch size} & {tokens}\\
            \midrule
            GPT2-Base 	&12	&768 & 1.0e-3& 0.5M & 9B  \\
            \midrule
            GPT2-Medium	&16	&1024 & 8.0e-4& 0.5M & 9B  \\
            \midrule
            GPT2-Large	&20	&1280 & 6.0e-4& 0.5M & 9B  \\
            \midrule
            GPT2-XL	    &25	&1600 & 4.0e-4& 0.5M & 9B  \\
            \bottomrule
            \end{tabular}}
            \label{tab:gpt2_hyper}
        \end{table}

    \subsection{Training Hyperparameters on Pythia}
    For the Pythia model family, we detail the training hyperparameters in Table~\ref{tab:pythia_hyper}. 
    Consistent with standard practices, the Pythia models are trained with a global batch size of 2M tokens for a total of 300B tokens. 

     \begin{table}[ht]
            \centering
            \caption{Training hyperparameters for Pythia family models. 
            }
            \scalebox{1}{
            \begin{tabular}{lccccc}
            \toprule
            Model & $n_{\mathrm{head}}$ & $d_{\mathrm{model}}$ & {learning rate} & {batch size} & {tokens}\\
            \midrule
            Pythia-70M 	&8	&512 & 1.0e-3& 2M & 300B  \\
            \midrule
            Pythia-160M	&12	&768 & 6.0e-4& 2M & 300B  \\
            \midrule
            Pythia-410M	&16	&1024 & 3.0e-4& 2M & 300B  \\
            \midrule
            Pythia-1B	&8	&2048 & 2.5e-4& 2M & 300B  \\
            \midrule
            Pythia-1.4B	&16	&2048 & 2.0e-4& 2M & 300B  \\
            \bottomrule
            \end{tabular}}
            \label{tab:pythia_hyper}
        \end{table}

    \subsection{Training Details of The Main Result}
    The primary computational overhead is incurred during the pretraining phase of ContextLM-Pythia-1B and ContextLM-Pythia-1.4B on the 300B-token Pile dataset. We conduct the pretraining on a high-performance computing cluster equipped with NVIDIA A100 GPUs (80GB VRAM). The total training compute amounts to approximately 3,594 and 5,375 GPU hours for the 1B and 1.4B models, respectively.

    \begin{table*}[ht]
        \centering
        \caption{Perplexity comparisons between GPT-2, GPT2-PM and ContextLM-GPT2 across four benchmark datasets. ContextLM consistently achieves lower perplexity across all model scales. }
        \renewcommand{\arraystretch}{1.3}
        \resizebox{1\textwidth}{!}{
        \begin{tabular}{lllll|l}
        \toprule
        Model & OWT & Wikitext & Lambada OpenAI & Lambada Standard & Avg PPL $\downarrow$ \\
        \midrule
        GPT2-Base & 22.38 & 45.41 & 74.06 & 301.82 & 110.92 \\
        GPT2-Base-PM & 21.61 & 43.17 & 66.84 & 299.46 & 107.77\\
        \rowcolor[HTML]{F2F3F5}
        \textbf{ContextLM-Base} & \textbf{20.68\textcolor{green!60!black}{\scriptsize{\({\downarrow1.70}\)}}} & \textbf{41.45\textcolor{green!60!black}{\scriptsize{\({\downarrow3.96}\)}}} & \textbf{55.22\textcolor{green!60!black}{\scriptsize{\({\downarrow18.84}\)}}} & \textbf{231.41\textcolor{green!60!black}{\scriptsize{\({\downarrow70.41}\)}}} & \textbf{87.19\textcolor{green!60!black}{\scriptsize{\({\downarrow 23.73}\)}}} \\
        \midrule
        GPT2-Medium & 18.10 & 35.78 & 36.53 & 131.43 & 55.46 \\
        GPT2-Medium-PM & 17.87 & 34.35 & 34.14 & 126.35 & 53.18 \\
        \rowcolor[HTML]{F2F3F5}
        \textbf{ContextLM-Medium} & \textbf{17.03\textcolor{green!60!black}{\scriptsize{\({\downarrow1.07}\)}}} & \textbf{32.08\textcolor{green!60!black}{\scriptsize{\({\downarrow3.70}\)}}} & \textbf{28.06\textcolor{green!60!black}{\scriptsize{\({\downarrow8.47}\)}}} & \textbf{95.06\textcolor{green!60!black}{\scriptsize{\({\downarrow36.37}\)}}} & \textbf{43.06\textcolor{green!60!black}{\scriptsize{\({\downarrow12.40}\)}}} \\
        \midrule
        GPT2-Large & 16.22 & 30.92 & 26.68 & 79.27 & 38.27  \\
        GPT2-Large-PM & 16.18 & 31.04 & 24.79 & 73.12 & 36.28\\
        \rowcolor[HTML]{F2F3F5}
        \textbf{ContextLM-Large} & \textbf{15.41\textcolor{green!60!black}{\scriptsize{\({\downarrow0.81}\)}}} & \textbf{28.82\textcolor{green!60!black}{\scriptsize{\({\downarrow2.10}\)}}} & \textbf{20.89\textcolor{green!60!black}{\scriptsize{\({\downarrow5.79}\)}}} & \textbf{62.09\textcolor{green!60!black}{\scriptsize{\({\downarrow17.18}\)}}} & \textbf{31.80\textcolor{green!60!black}{\scriptsize{\({\downarrow6.47}\)}}} \\
        \midrule
        GPT2-XL & 15.25 & 28.98 & 22.76 & 59.17 & 31.54  \\
        GPT2-XL-PM & 15.14 & 28.23 & 22.08 & 60.82 &  31.57 \\
        \rowcolor[HTML]{F2F3F5}
        \textbf{ContextLM-XL} & \textbf{14.60\textcolor{green!60!black}{\scriptsize{\({\downarrow0.65}\)}}} & \textbf{27.05\textcolor{green!60!black}{\scriptsize{\({\downarrow1.93}\)}}} & \textbf{17.46\textcolor{green!60!black}{\scriptsize{\({\downarrow5.30}\)}}} & \textbf{44.62\textcolor{green!60!black}{\scriptsize{\({\downarrow14.55}\)}}} & \textbf{25.93\textcolor{green!60!black}{\scriptsize{\({\downarrow5.61}\)}}} \\
        \bottomrule
        \end{tabular}
        }
        \label{tab:contextlm_results_ppl}
    \end{table*}

    \subsection{Training Curve Comparison}
    \label{sec:pythia_curve_comp}
\begin{figure*}[ht]
        \centering
        \includegraphics[width=0.6\linewidth]{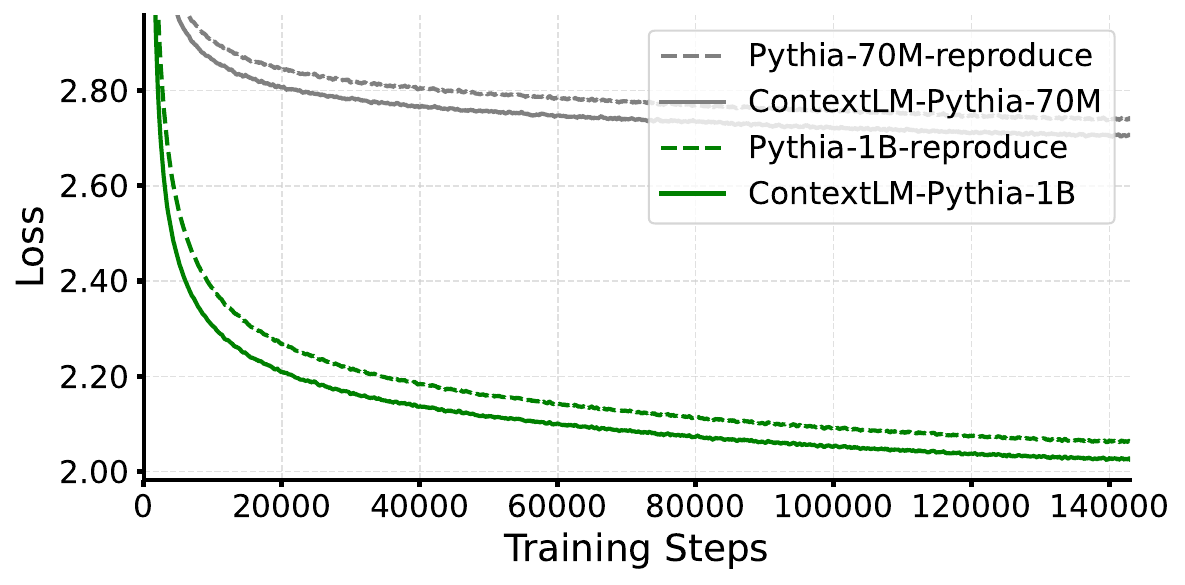}
        \color{blue} \caption{Training curve for Pythia vs ContextLM training on Pile (300B tokens). 
        Comparison 70M and 1B baselines to control for any differences in training configurations.
        }
        \label{fig:pythia_trainig_curve}
    \end{figure*}

    In this section, we compare the training loss convergence of ContextLM-Pythia against the Pythia baseline. Both models are trained from scratch on the Pile dataset (300B tokens) using the same data and hyperparameters. As Figure~\ref{fig:pythia_trainig_curve} demonstrated, ContextLM-Pythia consistently maintains a lower loss compared to the baseline, proving that the context prediction objective introduces a more efficient gradient signal that sustains long-term performance gains. 
    
        \subsection{Computational Complexity and Memory Footprint}
    \label{sec: computational complexity}
    We compare the computational complexity and memory requirements of \MethodName\hspace{1pt} with a parameter-matched baseline in Table~\ref{tab:complexity_analysis}. 
    \begin{itemize}[leftmargin=*]
        \item \textbf{Computational Complexity.} 
        The Token Decoder in \MethodName\hspace{1pt} shares the same $\mathcal{O}(T^2 d)$ attention complexity as the vanilla Transformer, where $T$ is the sequence length and $d$ is the hidden dimension. The Context Predictor operates on a chunked sequence of length $K$, resulting in a complexity of $\mathcal{O}(K^2 d) \approx \mathcal{O}(\frac{1}{w^2}T^2 d)$. 
    
        \item \textbf{Memory Footprint.} 
        The memory required for token embeddings is $\mathcal{O}(Td)$. The predicted context embeddings form a sequence of length $K$, contributing $\mathcal{O}((T/w)d)$ storage. 
    \end{itemize}
    \begin{table}[ht]
    \centering
    \caption{Complexity and efficiency analysis comparing GPT2-XL-PM and ContextLM-GPT2-XL on NVIDIA V100 GPUs.}
    \label{tab:complexity_analysis}
    \renewcommand{\arraystretch}{1.2}
    \setlength{\tabcolsep}{5pt} 
    \renewcommand{\arraystretch}{1.3}
        \resizebox{\textwidth}{!}{
    \begin{tabular}{lcccc}
    \toprule
    \multirow{2}{*}{\textbf{Component / Model}} & \multicolumn{2}{c}{\textbf{Theoretical Complexity}} & \multicolumn{2}{c}{\textbf{Practical Measurements}} \\
    \cmidrule(lr){2-3} \cmidrule(lr){4-5}
        & \textbf{Training} & \textbf{Inference} & \textbf{Train (tok/s)} $\uparrow$ & \textbf{Infer (ms/tok)} $\downarrow$ \\
    \midrule
    \rowcolor[HTML]{F2F3F5}
    \textit{Theoretical Components} & & & & \\
    \hspace{3mm}Token Decoder & $\mathcal{O}(T^2 d)$ & $\mathcal{O}(T d)$ & -- & -- \\
    \hspace{3mm}Context Predictor & $\mathcal{O}((T/w)^2 d)$ & $\mathcal{O}((T/w) d)$ & -- & -- \\
    \midrule
    \rowcolor[HTML]{F2F3F5}
    \textit{Practical Models} & & & & \\
    \hspace{3mm}GPT2-XL-PM  & $\mathcal{O}(T^2 d)$ & $\mathcal{O}(T d)$ & 1275.20 & 23.05 \\
    \hspace{3mm}ContextLM-GPT2-XL & $\approx \mathcal{O}((1 \!+\! 1/w^2)T^2 d)$ & $\approx \mathcal{O}((1 \!+\! 1/w)T d)$ & \textbf{1275.71} & \textbf{22.88} \\
    \bottomrule
    \end{tabular}}
    \end{table}

     \subsection{Downstream Task Evaluation on GPT-2}
    \label{appendix:downsream_task}
    To verify that ContextLM's improvements stem from the Context Predictor's modeling capabilities rather than simply the addition of parameters, we train an aggressively configured parameter-matched (GPT2-PM) baseline. The GPT2-PM model is constructed by adding standard Transformer layers to the vanilla GPT-2 backbone, explicitly aligning its total parameter count with that of ContextLM.

    As detailed in Table~\ref{tab:contextlm_gpt2_results_ppl}, our model achieves consistently lower perplexity than the PM baseline on all general language modeling benchmark. 
    We further extend the evaluation to $9$ representative benchmarks under both zero-shot and five-shot settings, as summarized in Table~\ref{tab:contextlm_results_acc_gpt2}. ContextLM demonstrates systematic improvements across nine representative benchmarks. These gains are consistent across all model scales, with particularly pronounced improvements on reasoning-intensive tasks such as HellaSwag and PIQA. 
    
    The stable performance enhancement across both perplexity-based and task-based metrics substantiates that context-level supervision strengthens generalization capability and promotes more robust compositional understanding across datasets and model sizes.
      \begin{table*}[ht]
        \centering
        \caption{Perplexity comparisons between GPT-2, GPT2-PM and ContextLM-GPT2 across four benchmark datasets. ContextLM consistently achieves lower perplexity across all model scales. }
        \renewcommand{\arraystretch}{1.3}
        \resizebox{\textwidth}{!}{
        \begin{tabular}{lllll|l}
        \toprule
        Model & OWT & Wikitext & Lambada OpenAI & Lambada Standard & Avg PPL $\downarrow$ \\
        \midrule
        GPT2-Base & 22.38 & 45.41 & 74.06 & 301.82 & 110.92 \\
        GPT2-Base-PM & 21.61 & 43.17 & 66.84 & 299.46 & 107.77\\
        \rowcolor[HTML]{F2F3F5}
        \textbf{ContextLM-Base} & \textbf{20.68\textcolor{green!60!black}{\scriptsize{\({\downarrow1.70}\)}}} & \textbf{41.45\textcolor{green!60!black}{\scriptsize{\({\downarrow3.96}\)}}} & \textbf{55.22\textcolor{green!60!black}{\scriptsize{\({\downarrow18.84}\)}}} & \textbf{231.41\textcolor{green!60!black}{\scriptsize{\({\downarrow70.41}\)}}} & \textbf{87.19\textcolor{green!60!black}{\scriptsize{\({\downarrow 23.73}\)}}} \\
        \midrule
        GPT2-Medium & 18.10 & 35.78 & 36.53 & 131.43 & 55.46 \\
        GPT2-Medium-PM & 17.87 & 34.35 & 34.14 & 126.35 & 53.18 \\
        \rowcolor[HTML]{F2F3F5}
        \textbf{ContextLM-Medium} & \textbf{17.03\textcolor{green!60!black}{\scriptsize{\({\downarrow1.07}\)}}} & \textbf{32.08\textcolor{green!60!black}{\scriptsize{\({\downarrow3.70}\)}}} & \textbf{28.06\textcolor{green!60!black}{\scriptsize{\({\downarrow8.47}\)}}} & \textbf{95.06\textcolor{green!60!black}{\scriptsize{\({\downarrow36.37}\)}}} & \textbf{43.06\textcolor{green!60!black}{\scriptsize{\({\downarrow12.40}\)}}} \\
        \midrule
        GPT2-Large & 16.22 & 30.92 & 26.68 & 79.27 & 38.27  \\
        GPT2-Large-PM & 16.18 & 31.04 & 24.79 & 73.12 & 36.28\\
        \rowcolor[HTML]{F2F3F5}
        \textbf{ContextLM-Large} & \textbf{15.41\textcolor{green!60!black}{\scriptsize{\({\downarrow0.81}\)}}} & \textbf{28.82\textcolor{green!60!black}{\scriptsize{\({\downarrow2.10}\)}}} & \textbf{20.89\textcolor{green!60!black}{\scriptsize{\({\downarrow5.79}\)}}} & \textbf{62.09\textcolor{green!60!black}{\scriptsize{\({\downarrow17.18}\)}}} & \textbf{31.80\textcolor{green!60!black}{\scriptsize{\({\downarrow6.47}\)}}} \\
        \midrule
        GPT2-XL & 15.25 & 28.98 & 22.76 & 59.17 & 31.54  \\
        GPT2-XL-PM & 15.14 & 28.23 & 22.08 & 60.82 &  31.57 \\
        \rowcolor[HTML]{F2F3F5}
        \textbf{ContextLM-XL} & \textbf{14.60\textcolor{green!60!black}{\scriptsize{\({\downarrow0.65}\)}}} & \textbf{27.05\textcolor{green!60!black}{\scriptsize{\({\downarrow1.93}\)}}} & \textbf{17.46\textcolor{green!60!black}{\scriptsize{\({\downarrow5.30}\)}}} & \textbf{44.62\textcolor{green!60!black}{\scriptsize{\({\downarrow14.55}\)}}} & \textbf{25.93\textcolor{green!60!black}{\scriptsize{\({\downarrow5.61}\)}}} \\
        \bottomrule
        \end{tabular}
        }
        \label{tab:contextlm_gpt2_results_ppl}
    \end{table*}
    
  \begin{table*}[ht]
    \centering
    \caption{Downstream task accuracy across nine benchmarks for GPT-2, GPT2-PM and ContextLM-GPT2 under zero-shot and five-shot settings. ContextLM-GPT2 consistently outperforms GPT-2 and GPT2-PM across all model scales.}
    \renewcommand{\arraystretch}{1.3}
    \resizebox{\textwidth}{!}{
    \begin{tabular}{lccccccccc|c}
    \toprule
    Model & \makecell{Lambada \\ OpenAI} & \makecell{ARC-\\E} & \makecell{Lambada \\ Standard} & \makecell{ARC-\\C} & \makecell{Wino\\Grande} & \makecell{PIQA} & \makecell{Hella-\\Swag} & \makecell{SciQ} & \makecell{RACE} & \makecell{Avg \\ Acc}$\uparrow$ \\
    \midrule
    \midrule
    \rowcolor[HTML]{E3E4E5}
    \multicolumn{11}{c}{\textit{\textbf{Zero-shot}}}\\
    \midrule
    GPT2-Base & 27.0 & 42.1 & 20.1 & 18.2 & 49.6 & 60.3 & 27.5 & 67.5 & 27.5 & 37.8 \\
    GPT2-Base-PM  & 27.3 & 41.2 & 19.8 & 18.9 & 50.3 & 60.2 & 27.6 & 69.2 & 27.2 & 38.0 \\
    \rowcolor[HTML]{F2F3F5}
    \textbf{ContextLM-Base} &  29.2 & 43.8 & 20.3 & 19.3 & 52.9 & 60.7 & 27.6 & 69.5 & 28.4 & \textbf{39.1 / \textcolor{green!60!black}{\textbf{\({\uparrow1.3}\)}}} \\
    \midrule
    GPT2-Medium & 33.3 & 44.1 & 23.9 & 19.3 & 50.5 & 62.9 & 29.0 & 72.7 & 29.1 & 40.5  \\
    GPT2-Medium-PM & 33.9 & 45.2 & 24.9 & 19.8 & 50.3 & 62.8 & 29.1 & 74.0 & 29.3 & 41.0  \\
    \rowcolor[HTML]{F2F3F5}
    \textbf{ContextLM-Medium} & 36.8 & 46.7 & 26.7 & 19.5 & 50.9 & 62.9 & 29.8 & 74.2 & 29.3 & \textbf{41.9 / \textcolor{green!60!black}{\textbf{\({\uparrow1.4}\)}}} \\
    \midrule
    GPT2-Large & 36.2 & 45.9 & 26.5 & 20.5 & 48.3 & 63.6 & 30.3 & 73.5 & 28.1 & 41.4  \\
    GPT2-Large-PM & 37.3 & 47.2 & 26.9 & 20.2 & 51.4 & 64.3 & 30.3 & 74.6 & 29.7 & 42.4  \\
    \rowcolor[HTML]{F2F3F5}
    \textbf{ContextLM-Large} & 40.7 & 47.9 & 29.9 & 21.1 & 51.3 & 65.7 & 31.2 & 76.6 & 31.0 & \textbf{43.9 / \textcolor{green!60!black}{\textbf{\({\uparrow2.5}\)}}}\\
    \midrule
    GPT2-XL & 38.7 & 48.1 & 29.2 & 20.6 & 50.8 & 65.2 & 31.0 & 77.3 & 29.2 & 43.3 \\
    GPT2-XL-PM  &  38.2 & 49.4 & 29.0 & 20.7 & 50.7 & 64.3 & 31.1 & 75.4 & 30.0 & 43.2 \\
    \rowcolor[HTML]{F2F3F5}
    \textbf{ContextLM-XL} & 41.9 & 49.9 & 32.3 & 21.4 & 51.9 & 66.1 & 32.3 & 77.7 & 31.2 & \textbf{45.0 / \textcolor{green!60!black}{\textbf{\({\uparrow1.7}\)}}}\\
    \midrule
    \midrule
    \rowcolor[HTML]{E3E4E5}
    \multicolumn{11}{c}{\textit{\textbf{Five-shot}}}\\
    \midrule
    GPT2-Base     & 18.3 & 40.8 & 17.0 & 18.6 & 51.9 & 59.8 & 27.3 & 69.0 & 26.8 & 36.6 \\
    GPT2-Base-PM  &18.8 & 40.5 & 16.7 & 18.9 & 50.4 & 60.3 & 28.0 & 69.4 & 27.4 & 36.7 \\
    \rowcolor[HTML]{F2F3F5}
    \textbf{ContextLM-Base} & 19.4 & 42.2 & 17.9 & 18.8 & 51.7 & 60.8 & 27.8 & 72.4 & 27.2 & \textbf{37.6 / \textcolor{green!60!black}{\textbf{\({\uparrow1.0}\)}}} \\
    \midrule
    GPT2-Medium    & 22.8 & 45.2 & 20.4 & 20.1 & 49.0 & 62.5 & 28.9 & 71.8 & 28.5 & 38.8 \\    
    GPT2-Medium-PM    & 22.7 & 44.7 & 21.6 & 21.4 & 48.7 & 61.3 & 29.2 & 75.4 & 30.0 & 39.4 \\
    \rowcolor[HTML]{F2F3F5}
    \textbf{ContextLM-Medium} & 24.0 & 47.3 & 22.0 & 20.1 & 51.1 & 64.2 & 29.8 & 75.3 & 28.4 & \textbf{40.2 / \textcolor{green!60!black}{\textbf{\({\uparrow1.4}\)}}} \\
    \midrule
    GPT2-Large  & 25.3 & 46.6 & 23.1 & 20.6 & 51.5 & 65.1 & 30.4 & 79.1 & 27.7 & 41.0 \\
    GPT2-Large-PM & 25.4 & 47.0 & 24.0 & 21.8 & 51.0 & 63.9 & 30.5 & 80.1 & 28.2 & 41.3 \\
    \rowcolor[HTML]{F2F3F5}
    \textbf{ContextLM-Large} & 29.7 & 49.7 & 26.4 & 21.1 & 51.5 & 65.2 & 31.2 & 82.0 & 30.7 & \textbf{43.1 / \textcolor{green!60!black}{\textbf{\({\uparrow2.1}\)}}} \\
    \midrule
    GPT2-XL   & 28.0 & 49.0 & 24.3 & 21.2 & 50.6 & 65.4 & 30.9 & 81.3 & 28.4 & 42.1  \\
    GPT2-XL-PM & 27.5 & 50.5 & 25.1 & 21.3 & 50.9 & 64.9 & 31.3 & 80.4 & 29.3 & 42.4 \\
    \rowcolor[HTML]{F2F3F5}
    \textbf{ContextLM-XL} & 30.7 & 50.6 & 29.3 & 22.2 & 51.0 & 65.9 & 32.2 & 83.6 & 30.8 & \textbf{44.0 / \textcolor{green!60!black}{\textbf{\({\uparrow1.9}\)}}} \\
    \bottomrule
    \end{tabular}
    }
    \label{tab:contextlm_results_acc_gpt2}
    \end{table*}

    \subsection{Ablation of Aggregation Strategy} 
    We evaluate different aggregation mechanisms using Pythia-ContextLM-410M at the 30k-step checkpoint. As shown in Table~\ref{tab:pooling_ablation}, the performance differences across different aggregation strategies are marginal. Consequently, we select Mean Pooling as our default design, as it achieves competitive performance comparable to more complex parameterized methods without introducing any additional parameters.

    \begin{table}[h]
    \centering
    \caption{Ablation of Aggregation Strategies on Pythia-ContextLM-410M at the 30k-step checkpoint. 
    }
    \label{tab:pooling_ablation}
    \begin{tabular}{lc}
    \toprule
    \textbf{Strategy} & \textbf{Loss} $\downarrow$ \\
    \midrule
    \textbf{Mean Pooling (Ours)} & \textbf{2.2919} \\
    MLP Pooling & 2.2930\\
    Latent Attention Pooling & 2.2953 \\
    Self-Attention Pooling & 2.2907 \\
    \bottomrule
    \end{tabular}
    \end{table}

\end{document}